\documentclass{article}
\usepackage[utf8]{inputenc}
\usepackage[fleqn]{amsmath}
\usepackage{longtable}
\usepackage{graphicx}
\usepackage{booktabs}
\usepackage{float}
\usepackage{colortbl}
\usepackage{multirow}
\usepackage{rotating}
\usepackage{xcolor}
\usepackage[ruled,vlined]{algorithm2e}
\usepackage{tabu}
\usepackage{algpseudocode} 
\usepackage{amssymb}
\usepackage{amsmath}
\usepackage[normalem]{ulem}
\usepackage{apacite}
\usepackage{natbib}
\usepackage{graphicx}
\usepackage{authblk}
\usepackage{microtype}
\usepackage[breaklinks=true]{hyperref}
\usepackage{breakcites}

\title{Dynamic Ensemble Learning for Credit Scoring: A Comparative Study}
\date{}

\begin{document}

\author[1]{Mahsan Abdoli \thanks{\texttt{m.abdoli@aut.ac.ir} (Mahsan Abdoli)}}
\author[2]{Mohammad Akbari \thanks{Corresponding author: \texttt{akbari.ma@aut.ac.ir} (Mohammad Akbari)}}
\author[1]{Jamal Shahrabi \thanks{\texttt{jamalshahrabi@aut.ac.ir} (Jamal Shahrabi)}}

\affil[1]{Department of Industrial Engineering, Amirkabir University of Technology (Tehran Polytechnic)}
\affil[2]{Department of Mathematics and Computer Science, Amirkabir University of Technology (Tehran Polytechnic)}

\maketitle

\section{Abstract}
Automatic credit scoring, which assesses the probability of default by loan applicants, plays a vital role in peer-to-peer lending platforms to reduce the risk of lenders. Although it has been demonstrated that dynamic selection techniques are effective for classification tasks, the performance of these techniques for credit scoring has not yet been determined. This study attempts to benchmark different dynamic selection approaches systematically for ensemble learning models to accurately estimate the credit scoring task on a large and high-dimensional real-life credit scoring data set. The results of this study indicate that dynamic selection techniques are able to boost the performance of ensemble models, especially in imbalanced training environments. 
\\ \textbf{Keywords:} Credit Scoring, Peer-to-peer Lending Platform, Dynamic Ensemble Learning

\section{Introduction}

The need for credit scoring goes back to the beginning of borrowing and lending. Lenders often attempt to gather information about loan applicants to distinguish reliable customers from unreliable ones based on the likelihood of being charged-off \citep{louzada2016classification}.
The objective of credit scoring is to assess the probability that a borrower will show some undesirable behavior in the future.
Financial institutions leverage scorecards, which are predictive models developed by classification algorithms to estimate the probability of default (PD) by loan applicants \citep{lessmann2015benchmarking}.

With the emergence of digital technology, social lending, also known as peer-to-peer (P2P) lending, becomes an alternative to the traditional loan granting process in which individuals lend and borrow money through an online platform that connects borrowers to lenders without the intermediation of banks. P2P lending platforms such as  Prosper~\footnote{https://www.prosper.com/}, Lending Club~\footnote{https://www.lendingclub.com/}, and Zopa~\footnote{https://www.zopa.com/} ~\citep{malekipirbazari2015risk} can substantially reduce intermediary costs compared to traditional banks due to lack of a "brick-and-mortar" business approach \citep{teply2019best}.

P2P lending can reduce the processing time and cost for both borrowers and lenders. Borrowers can request micro-loans directly from lenders with a lower interest rate and faster processing time. Lenders can earn higher rates of return with fewer administrative fees compared to traditional savings accounts. In traditional lending markets, banks and financial institutions can use collateral as a tool to enhance lenders' trust in borrowers. Such actions to increase trust between borrowers and lenders cannot be implemented in an online environment. In such a context, automated scoring of customers plays a vital role in credit risk assessment~\citep{emekter2015evaluating}.

With the expansion of financial institutions' loan portfolios as well as the advent of P2P lending platforms, classifying customers based on their PD is crucial to support decision making. In credit scoring literature, it is well established that a minor improvement in the credit scoring accuracy can result in significant future savings~\citep{baesens2003benchmarking}. 
As such, various credit scoring models are exploited by banks, financial institutions, and online P2P lending platforms to make informed decisions regarding the borrowers' default risk. Thus, available customer data, such as loan history, demographic information, financial and educational data, is exploited to build a machine learning model which then is used to develop a decision support system to group loan applications into reliable and unreliable ones.

Despite the significance and importance of credit scoring in reducing risks and costs of lenders and financial institutions, the performance of credit scoring models is not satisfying yet for practical applications in real life due to the following issues. First, the number of available datasets for credit scoring is limited due to the difficulty of obtaining customers' credit data. Therefore, credit scoring studies use different public and private data sets. The public data sets used in credit scoring are often small or contain encrypted variables due to privacy concerns. On the other hand, private data sets cannot be released to the public \citep{louzada2016classification}. In such a context, new classification approaches are evaluated in different datasets and/or settings. The effectiveness of different approaches is unclear due to the lack of a systematic benchmark of various credit scoring models. Hence, the comparison of models trained on different data sets is highly desired.

Second,  several statistical techniques and machine learning approaches have been proposed over the years to improve classification performance of credit scoring such as Linear Discriminant Analysis (LDA) \citep{reichert1983examination}, Logistic Regression (LR) \citep{hand2002superscorecards}, Support Vector Machine (SVM) \citep{huang2004credit}, Neural Networks (NN) \citep{bastani2019wide,duan2019financial}, etc. In addition to individual classifiers, lately, there has been much attention to ensemble models that aggregate the classification power of individual classifiers (base classifiers) to improve the final result such as \citep{xia2018novel,he2018novel,yu2018dbn, ala2016classifiers,ala2016new}. One of the challenges of implementing credit scoring models in real-life is that, despite the wide range of proposed classification techniques in credit scoring literature, the most effective techniques for different data sets, especially real-life data sets, have not yet been determined. Therefore, future studies should focus on comparing the abilities of different classification approaches.  

Finally, similar to many other real-life classification problems, credit scoring data sets are heavily imbalanced \citep{brown2012experimental, xiao2012dynamic}. The majority of samples in the credit scoring data sets belong to the negative class (i.e., the loans that were fully paid). Therefore, robust algorithms that can handle imbalanced data should be developed to increase the classification performance.

While several machine learning approaches have been employed to improve the performance of credit scoring models, recently, \citet{lessmann2015benchmarking} have demonstrated that multiple classifier system (also known as ensemble models) are able to outperform individual classifier.

Based on the idea that building a single classifier to cover all the intrinsic variabilities of the data sets cannot take advantage of all the available information in the data set, Multiple Classifier Systems (MCSs) were introduced. MCSs use multiple classifiers' decisions to make a more robust and effective model for predicting the class of a sample \citep{britto2014dynamic}. The MCSs can be divided into two categories: Static Selection (SS) and Dynamic Selection (DS). In static selection, the strategy to select the best base classifiers is determined in the training set, which is then applied on all test samples, regardless of the competency of the base classifier in the local region surrounding the test sample. In dynamic selection, the most competent classifiers in the local region of the test sample are selected based on a competence criterion for each test sample on the fly. Therefore, each test sample is classified by one or a set of classifiers that have a high performance in the local region of the test sample based on the competency criterion used for the base classifier selection. 
The rationale for dynamic selection strategies is that each base classifier in the pool of classifiers is an expert in a specific local region of the feature space. Therefore, the most competent classifiers should be selected to classify each test sample from the pool of classifiers \citep{cruz2018dynamic}. 

The effectiveness of using dynamic selection in classification task has been tested by \citet{britto2014dynamic} and \citet{cruz2018dynamic}; In their studies, the performance of DS was evaluated on several benchmarking data sets. \citet{britto2014dynamic} and \citet{cruz2018dynamic} concluded that using DS can boost the performance of a pool of weak classifiers. \citet{cruz2018dynamic} state that as dynamic selection techniques perform locally, the final classification results are not biased towards the majority class, which will be validated in this study experimentally. 
\citet{lessmann2015benchmarking} used two dynamic selection algorithms in their benchmark to evaluate their performance on credit scoring data sets. \citet{junior2020novel} used four dynamic selection techniques to evaluate the performance of credit scoring. However, to the best of our knowledge, no comprehensive study has been conducted to evaluate other dynamic selection techniques on the credit scoring problem specifically on a real world data set. 

In this study, four classifiers, namely support vector machine (SVM), multilayer perceptron (MLP), k nearest neighbors (k-NN), and Gaussian naive Bayes (GNB) are used to construct the pool of classifiers. In addition to individual classifiers, DS techniques are also applied to the random forest (RF) to evaluate their effectiveness to boost the performance of classification. The classifiers are evaluated on the Lending Club data set, which is a real-world data set in the field of credit scoring and social lending. Furthermore, the ability of dynamic selection to classify test samples trained on data sets with different imbalance ratios is evaluated. Five data sets with different imbalance ratios created by under-sampling the majority class, as well as the original data set, are used to investigate the impact of imbalanced data on the robustness of classification performance.

\section{Literature Review}

\citet{louzada2016classification} present a systematic review of 187 papers from 1992 to 2015 on binary classification techniques for the credit scoring problem. The papers in this systematic review are categorized based on seven objectives: proposing a new method, comparing traditional techniques, conceptual discussions, feature selection, literature review, performance measures studies, and others.

With regards to the categories mentioned in the review of \citet{louzada2016classification} many new DS algorithms have been developed to classify default of borrowers. For example, \citet{xiao2016ensemble} use supervised clustering to make different ensembles in each cluster. By assigning each test sample to the cluster with the most similar members they use different sets of classifiers to classify each test sample. Unsupervised clustering integrated with fuzzy assignment is implemented in \citet{zhang2018classifier} work in the pool generation and testing phase of the classification procedure. They also use genetic algorithm to include both diversity and accuracy measures in selecting the best classifiers in the field of credit scoring. \citet{xiao2012dynamic} combine ensemble learning method with cost-sensitive learning for imbalanced data sets  to develop a framework that for each test sample selects the most appropriate classifier(s) to classify customers. \citet{feng2019dynamic} built a dynamic weighted trainable combiner based on Markov Chain to model the impact of selecting different classifiers in decreasing the missclassification cost in credit scoring. 

Regarding the  comparison of different classification techniques for credit scoring, several studies have been carried out: \citet{baesens2003benchmarking} have studied the performance of various classification algorithms such as logistic regression, linear and quadratic discriminant analysis, support vector machines, neural networks, naive bayes and nearest neighbour classifiers. Among the applied classifiers, least squares support vector machines and neural networks yielded a good performance on several credit scoring data sets. Following that, \citet{lessmann2015benchmarking} have provided a benchmark of 41 classification algorithms across credit scoring data sets. In addition to individual classifiers covered by \citet{baesens2003benchmarking},  homogeneous and heterogeneous ensemble models were also included.

As credit scoring data sets are usually highly imbalanced, a comparison of different classifiers applied to five credit scoring data sets was presented by \citet{brown2012experimental}. Logistic regression, least square support vector machine, decision trees, neural networks, k-NN, gradient boosting algorithm and random forests were applied while the imbalance ratio of the data set is increased by undersampling minority classes to evaluate the robustness of classifiers in extreme imbalanced situations. 

Apart from studies that focus on only comparing different classification models for credit scoring, papers that propose new classification techniques such as \citep{xiao2012dynamic, ala2016classifiers,ala2016new, xiao2016ensemble}, compare their methods with basic individual and ensemble methods. The standard procedure in the comparison process of new classification techniques is to compare them with LR, which is considered the industry standard. \citet{lessmann2015benchmarking} argue that LR should not be considered the baseline method in credit scoring because it may lead to a biased decision about the performance of newly proposed models, as random forest performs much better on credit scoring data sets as a simple algorithm.

Considering the comparative studies on the Lending Club data set, \citet{teply2019best} have conducted a comparison study to rank 10 different classifiers to evaluate the performance of different classifiers. According to the ranking, they have concluded that logistic regression and artificial neural networks were placed as the best and second-best classifiers respectively. They have also concluded that classification and regression trees and k-nearest neighbors perform poorly on the Lending club data set.

By comparing RF, SVM, LR, and k-NN, \citet{malekipirbazari2015risk} have shown that RF outperforms other classification methods and can be used as a powerful approach to predict borrowers' status.
\citet{junior2020novel} proposed a modification of k-NN algorithm to be used in the definition of local region in Dynamic selection techniques. In their study they have compared four dynamic selection techniques, namely Local Class Accuracy, Modified Classifier Rank, K-Nearest Oracles Eliminate, and K-Nearest Oracles Union on credit scoring data sets with different pre-processing and pool generation settings. \citet{junior2020novel} have concluded that K-Nearest Oracles Union could achieve the best result compared to the other three DS techniques in credit scoring.

As stated in \citet{cruz2018dynamic}, Dynamic selection is an active field of study in machine learning. So far, to the best of our knowledge only  \citet{lessmann2015benchmarking}, \citet{xiao2012dynamic}, and \citet{junior2020novel} have applied a limited number of dynamic  selection techniques on credit scoring data sets. Also as stated in \citet{junior2020novel}, we believe that examining the ability of DS techiniques, especially on real-world data sets to evaluate their competency is yet to be studied. In this paper, we apply 14 different DS techniques on different classification algorithms both in the context of homogeneous and heterogeneous pool of classifiers across varying degrees of data imbalance.

\section{methodology}
\subsection{Multiple classifier systems}
To deal with uncertainty and noise in data, various classification techniques have been developed over the years to address the limitations and improve classification performance. Due to the intrinsic characteristics of different classification models, the misclassified samples by different classifiers do not necessarily overlap. Therefore, different classification models potentially offer complementary information for classification of different test samples. Thus, fusing multiple classifiers often result in improvement of classification performance since each classifier provides complementary information for distinct aspects of a given sample \citep{kittler1998combining}. Therefore, MCS decisions are expected to improve the classification accuracy by combining the decisions of different classifiers trained on the training set \citep{dietterich2000experimental}.

Multiple classifier systems are composed of three phases: I) pool generation II) selection III) combination. In the first phase, a pool of accurate and diverse classifiers is constructed to classify the samples. The need for diverse classifiers comes from the fact that the generated classifiers should show some degree of complementarity. Bagging \citep{breiman1996bagging}, boosting \citep{freund1996experiments}, and random subspace \citep{ho1998random} are among the most commonly used strategies to create the pool of classifiers.

In the second phase, which is an optional phase, the generated classifiers are selected based on a competency measure to classify the unknown samples. There are two types of base classifier selection in the second phase: static and dynamic selection.   In static selection, the classifiers’ competence is determined in the training phase by computing the base classifiers' competency based on a selection criterion. After the classifier selection, all the selected base classifiers are employed to classify unknown samples regardless of the individual characteristics of the query sample in the selection of base classifiers. 

In dynamic selection, a single or a subset of trained classifiers are selected to classify each unknown sample exclusively regarding its surrounding local region. Based on the number of classifiers selected in DS techniques, they are categorized into two categories: I) Dynamic Classifier Selection (DCS), which selects only the most competent classifier from the pool of classifiers. II) Dynamic Ensemble Selection (DES) which selects a subset of classifiers from the pool.

The third phase of MCSs deals with the aggregation of decisions made by the selected classifies.  The outputs of the classifiers are aggregated according to a combination rule. One of the most basic combination rules that can be named is majority voting (i.e., aggregating the predictions of base classifiers and choosing the prediction with the most votes).

\subsection{Dynamic Selection}
In dynamic selection, the classification of an unknown sample is composed of the following steps:
First, the dynamic selection set (DSEL), which is a set of labeled samples from the training or validation set, is set aside to determine the region of competence. Here, the region of competence is the local region surrounding the query test sample, which is determined by the most similar or nearest samples from the DSEL. 
A test sample’s region of competence can be obtained by applying K-nearest neighbors technique, clustering and the competence map to the DSEL. The k nearest technique and clustering methods find the nearest and the most similar samples in the DSEL to the query sample which are then used to evaluate the competency of base classifiers. The competence map uses all samples in the DSEL as the region of competence. Then by applying Gaussian potential function, the influence of each DSEL sample in competency of classifers is computed.

Second, the selection criteria to compute the competency of each classifier in the region of competence is determined. These criteria can be calculated by the accuracy of the base classifier, their rank among all the classifiers present in the pool, or probabilistic methods.

Finally, one or a subset of classifiers from the pool are selected based on the competency level of classifiers in the region of competence. To classify the query sample, the final classification result is reached by combining chosen competent classifiers using a voting combination method such as majority voting.

For a comprehensive explanation and review of dynamic classifier selection techniques, we refer to \citet{cruz2018dynamic} and \citet{britto2014dynamic}. Before diving into details, we introduce the mathematical notation used in this paper as summarized in Table \ref{tab:notation_table}: 

\begingroup
\renewcommand{\arraystretch}{1.35}
\setlength{\tabcolsep}{6 pt}
\begin{table}[H]
\centering
\caption{The mathematical notation used in this study}
\label{tab:notation_table}
\resizebox{\linewidth}{!}{%
\begin{tabular}{ll} 
\toprule
Notation & Description \\ 
\hline
$C=\{c_1,\dots,c_M\}$  & The pool of classifiers with $M$ base classifiers.  \\
$x_j$  & The test sample to be classified. \\
$\theta_j=\{x_1,\dots,x_k\}$  & The region of competence surrounding $x_j$, $x_k$ is one sample belonging to the region of competence.  \\
$\Omega=\{\omega_1,\dots,\omega_L\}$  & The set of $L$ classes in the classification problem.  \\
$\omega_l$  & The class predicted by classifier $c_i$.  \\
$P(\omega_l | x_j, c_i)$  & Posterior probability of classifying $x_j$ as $\omega_l$ by classifier $c_i$.  \\
$W_k=\frac{1}{d_k}$  & The weight of $x_k$ computed by its distance from $x_j$, $d_k$ is the distance between the query $x_j$ and $x_k$.  \\
$\delta_i,_j$  & The estimated competency of classifier $c_i$ in classification of query $x_j$.  \\
$\tilde{x_j}$  & The output profile of the query sample $x_j$.  \\
$\phi_j$  & The set of most similar output profiles of the query sample, $\tilde{x_j}$, computed in the decision space.  \\
\bottomrule
\end{tabular}
}
\end{table}
\endgroup

\subsection{Dynamic selection techniques}
In this section, we briefly introduce various dynamic selection techniques:
\subsubsection{Modified Classifier Ranking}
In the modified classifier ranking technique by \citet{woods1997combination} and \citet{sabourin1993classifier}, the local accuracy of classifier $c_i$ is estimated by the number of consecutive correctly classified samples in the region of competence of a given test sample. The number of correctly classified samples is considered the rank of the classifier. The classifier with the highest rank in this technique is selected as the most competent classifier to classify the test sample.

\subsubsection{Overall Local Accuracy}
Overall locan accuracy (OLA)  calculates the competency of base classifier $c_i$ by computing the percentage of samples in the region of competence that have been correctly classified as shown in equation \ref{OLA} \citep{woods1997combination}.

\begin{flalign} \label{OLA}
\delta_i,_j =\frac{1}{K}\sum_{k=1}^{K} P(\omega_l | x_k\in \omega_l,c_i)
\end{flalign}

\subsubsection{Local Classifier Accuracy}
Local classifier accuracy (LCA) is proposed as the percentage of the samples in the region of competence that are correctly labeled by classifier $c_i$ with respect to the output class $\omega_l$. $\omega_l$ is the class with maximum probability assigned by classifier $c_i$. Equation \ref{LCA} represents the competence of classifier $i$ in the classification of query sample $x_j$ \citep{woods1997combination}.

\begin{flalign}\label{LCA}
\delta_i,_j =\frac{\sum_{x_k \in \omega_l} P(\omega_l|x_k,c_i)}{\sum_{k=1}^{K} P(\omega_l|x_k,c_i)}
\end{flalign}

\subsubsection{A Priori}
As presented in the equation \ref{apriori}, the selection criterion of A Priori technique is the probability of correctly classified samples in the region of competence by classifier $c_i$ \citep{giacinto1999methods}. Here, the probabilities are weighted by the Euclidean distance of $x_k$ from the query sample $x_j$. The classifier with the highest value of competency is selected if the competence level is significantly better than other base classifiers in the pool. If the competence level is not significantly higher for any of the classifiers, all classifiers in the pool are combined by majority voting.

\begin{flalign}\label{apriori}
\delta_i,_j =\frac{\sum_{k=1}^{K} P(\omega_l|x_k \in \omega_l,c_i)W_k}{\sum_{k=1}^{K} W_k}
\end{flalign}

\subsubsection{A Posteriori}
A Posteriori technique takes into account $\omega_l$, which is the assigned class to the query sample $x_j$, by classifier $c_i$ as shown in equation \ref{aposteriori} \citep{giacinto1999methods}. Similar to the A Priori technique, the classifier with a significantly higher competency level is chosen to classify the test sample. Otherwise, all classifiers in the pool are combined by majority voting to classify the test sample.

\begin{flalign}\label{aposteriori}
\delta_i,_j =\frac{\sum_{x_k \in \omega_l} P(\omega_l|x_k,c_i)W_k}{\sum_{k=1}^{K} P(\omega_l|x_k,c_i)W_k}
\end{flalign}

\subsubsection{Multiple Classifier Behavior}
A behavior knowledge space (BKS) is a k-dimensional space where each dimension corresponds to the decision of one base classifier    \citep{huang1995method}. Based on the behavior knowledge space and the classifier local accuracy, the multiple classifier behavior (MCB) technique computes the output profiles of each test sample (i.e., predicted class labels) along with the output profiles of the samples in the region of competence. The similarities between the output profile of a given test sample and the output profiles of instances in the region of competence, are calculated by $S(\tilde{x_j},\tilde{x_k})$, i.e., the similarity measure between $x_j$ and $x_k$ as demonstrated in equation \ref{MCB} and equation \ref{MCB_2}. Samples with the $S(\tilde{x_j},\tilde{x_k}) > \zeta$ are selected to remain in the region of competence and other samples in the region are deleted. Here, $\zeta$ is a hyper-parameter of the model which demonstrates the similarity threshold. The competency of each classifier in the pool is calculated by the classification accuracy in the modified region of competence. The classifier with significantly higher performance in the region of competence than other classifiers, is selected to classify $x_j$. If no such classifier exists, all classifiers in the pool of classifiers are combined using majority voting \citep{giacinto2001dynamic}. 

\begin{flalign}\label{MCB}
\text{S(}\tilde{x_j},\tilde{x_k})=\frac{1}{M}\sum_{i=1}^{M} T(x_j,x_k)
\end{flalign}

\begin{flalign}\label{MCB_2}
T(x_j,x_k)=
\begin{cases}
  \text{1} \qquad \text{if} \qquad c_i(x_j) = c_i(x_k),\\
  \text{0} \qquad \text{if} \qquad c_i(x_j) \neq c_i(x_k).
\end{cases}
\end{flalign}

\subsubsection{Modified Local Accuracy}
The hypothesis behind local accuracy is that the neighboring elements in the feature space maintain a stronger relationship with each other compared to further elements. Therefore weighting the impact of samples in the region of competence is logical. \citet{smits2002multiple} proposed a modified local accuracy (MLA) by weighting $x_k$ in the region of competence, using its distance to the query $x_j$, as shown in equation~\ref{MLA}. The classifier with the highest competence level is selected to classify the test sample.

\begin{flalign}\label{MLA}
\delta_i,_j =\sum_{k=1}^K P(\omega_l|x_k \in \omega_l, c_i)W_k
\end{flalign}

\subsubsection{DES-Clustering}
In DES-Clustering technique, the k-means algorithm is exploited to group DSEL into k distinct clusters. For each cluster, the classifiers are ranked in descending order of accuracy and ascending order of the double-fault diversity measure which is a pairwise diversity measure and uses the proportion of the missclassified samples by two classifiers. The test sample is assigned to a cluster by measuring its Euclidean distance to the cluster centroid. Finally N most accurate and J most diverse classifiers (N $\geq$ J) are selected to classify the test sample by combining the results of each classifier for the test sample \citep{soares2006using}.

\subsubsection{DES-KNN}
Here, the region of competence is determined by applying the k nearest neighbors method on the DSEL. Then, the classifiers in the pool $C=\{c_1,....,c_M\}$, are ranked decreasingly based on accuracy and increasingly based on double-fault diversity measure. The first N classifiers according to accuracy rank and the first J classifiers according to diversity rank (N $\geq$ J) are selected to classify the test sample \citep{soares2006using}. It is worth noting that DES-KNN is different from DES-Clustering with regards to the technique used in determining the region of competence for the given test sample $x_j$.

\subsubsection{K-Nearest Oracles Eliminate}
The Oracle is an abstract model that always selects the classifier that predicts the correct label for a given query if it exists \citep{kuncheva2002theoretical}. It is obvious that, the Oracle is regarded as a possible upper limit for MCSs performance in classification. This concept is adopted in K-Nearest Oracles Eliminate (knoraE). K-Nearest oracles Eliminate selects all the classifiers that classify all samples in the region of competence correctly if such classifier exists \citep{ko2008dynamic}. If no classifier can be found that satisfies the above criterion, the size of region of competence is reduced and the search to find the competent classifiers restarts.

\subsubsection{K-Nearest Oracles Union}
The classifiers that can classify at least one sample in the region of competence are selected in K-Nearest Oracles Union (knoraU) technique. The number of votes classifier $c_i$ has in the voting process is equivalent to the number of correctly classified samples in the region of competence to promote classifiers that participate more in making the correct decision (i.e., assigning the correct label to the query sample). In this manner, an automatic weighting procedure is done in the majority voting process \citep{ko2008dynamic}.

\subsubsection{Dynamic Ensemble Selection Performance}
\citet{woloszynski2012measure} introduced the dynamic ensemble selection performance (DESP), which attempts to compute the performance of $c_i$ in the region of competence . The competency of classifier $c_i$ is computed by the difference between the accuracy of $c_i$ in the region of competence and the accuracy of a random classifier in the region. The random classifier is presented as a classifier that randomly assigns classes to a sample with equal probabilities. Therefore, its performance in the region is $1/L$. Equation \ref{desp} shows the selection criterion of this technique. The base classifiers that are able to classify samples with a higher accuracy than the random classifier are selected to classify $x_j$.

\begin{flalign}\label{desp}
\delta_i,_j = \hat{P}(c_i|\theta_j)-\frac{1}{L}
\end{flalign}

\subsubsection{k-nearest Output Profiles}
By applying classifier $c_i$ to the test sample, and the DSEL set, the output profiles of the samples are obtained. Next, the similarity degree between the query output profile, $\tilde{x_j}$, and the output profiles of the DSEL samples is computed. Finally K most similar DSEL samples are selected as the region of competence. The classifiers that can correctly classify the region of competence samples are selected to classify the test sample in k-nearest output profiles (KNOP) technique. Similar to knoraU, each classifier gains one vote for each correct classification of samples in the region of competence \citep{cavalin2012logid}.

\subsubsection{Meta-DES}
A classifier’s level of competence determines its selection to classify a test sample in DS. In the DS literature, several competence measures have been proposed. To benefit from the advantages of different competency measures a meta-learning problem was generated by \citet{cruz2015meta} as follows:

\begin{itemize}
    \item The meta-classes are either "competent" (1) or “incompetent" (0) to classify $x_j$.
    \item Each set of meta-features $f_i$ corresponds to a different criterion for measuring the competency level of a base classifier.
    \item The meta-features are encoded into a meta-features vector $v_i,_j$.
    \item A meta-classifier $\lambda$ is trained based on the meta-features to predict whether or not the classifier $c_i$ will correctly classify $x_j$. In other words, if classifier $c_i$ is competent to classify the test sample.
\end{itemize}
In this technique, a meta classifier is trained to predict whether a base classifier is competent to classify a test sample.  In the meta training stage, the meta-features are extracted from the samples in the training and DSEL sets. The extracted meta-features are then used to train a meta classifier to classify each model in the pool of classifiers as competent or incompetent. The training phase of the Meta-DES technique is represented in Algorithm \ref{metades alg}.
In the generalization stage, meta-features, which are the meta classifier's input, are extracted for query $x_j$. The classifier $c_i$ is then evaluated by the meta classifier to determine whether it should be selected to participate in the classification of the given test sample or not.

\begin{algorithm}[ht]
\SetAlgoLined
\KwIn{$T_{\lambda}$: Training data}
\KwOut{Pool of Classifiers $C=\{c_1,\dots,c_M\}$}
 $T_{\lambda}^{*} = \varnothing$\;
 \For{$x_{n,train_{\lambda}} \in T_{\lambda}$}{
 Compute the consensus of the pool $H(x_{n,train_{\lambda}}, C)$ \\
 \uIf{$H(x_{n,train_{\lambda}}, C) < h_{c}$}{
 Find the region of competence $\theta_{j}$ of $x_{n,train_{\lambda}}$ using $T_{\lambda}$ \\
 Compute the output profile ${\Tilde{x}_{n,train_{\lambda}}}$ of $x_{n,train_{\lambda}}$ \\
 Find the $K_p$ similar output profiles $\phi_{j}$ of $x_{n,train_{\lambda}}$ \\
 \For{$c_i \in C$}{
 $v_{i,n}$= \textit{MetaFeatureExtraction($\theta_j,\phi_j,c_i,x_{n,train_{\lambda}}$) \\
 \uIf{$c_i$ correctly classifies $x_{n,train_{\lambda}}$}{$\alpha_{i,j}=1$ "$c_i$ is competent to classify $x_{n,train_{\lambda}}$" \\}
 \uElse{$\alpha_{i,j}=0$ "$c_i$ is incompetent to classify $x_{n,train_{\lambda}}$"}
 $T_{\lambda}^{*} = T_{\lambda}^{*}\bigcup \{v_{i,n}\}$ \\}
 }
 }
 }
 Divide $T_{\lambda}^{*}$ into 25\% for validation and 75\% for training \\
 Train $\lambda$ \\
 \Return{The meta-classifier $\lambda$}
 \caption{The META-DES Algorithm (Training Phase)}
 \label{metades alg}
\end{algorithm}

\section{Experimental Setting}
\subsection{Data set}
In this study we used a real-life data set from Lending Club~\footnote{retrieved from \url{https://-www.kaggle.com/wordsforthewise/lending-club}} which is one of the most popular P2P lending platforms in the financial industry \citep{malekipirbazari2015risk}. In order to apply for a loan, Lending Club users send their request via the website and the investors interested in investing, start funding for the requested loans based on the available information on the loan requests. 

The payback duration of loans provided by Lending Club is either $36$ or $60$ months. As the latest payment date made by a borrower was March 2019, we selected 36 month loans issued before March 2016 to ensure that loans have had enough time to reach their maturity. Out of the $151$ available variables in the original data set, we dropped variables that had more than half the size of the data set missing values. After considering samples labeled as “Charged off” and “Fully paid” as class 1 (bad customers) and class 0 (good customers), respectively we organized the remaining variables based on other studies that have used the Lending Club data set \citep{setiawan2019comparison,teply2019best, serrano2015determinants,malekipirbazari2015risk} and imputed the missing values of continuous variables by mean imputation. 

We stored the average value of variables \textit{fico\_range\_low} and \textit{fico\_range\_high} in variable \textit{average\_fico} for each data sample. We normalized continuous variable to be in range $[0, 1]$ and as a last step in data preparation we used dummy variables to represent values of categorical variables. The final variables and their description are shown in Table~\ref{Variables table}. 

We selected Loans issued from July 2015 to December 2015 as the training data set, the remainder as test set. The training set icludes $162,235$ samples (approximately $76\%$ of total data), and the test set contains $50,695$ samples (approximately $24\%$ of total data) from January and February of 2016. 

\begin{longtable}{@{}p{110.0 pt}p{223.0 pt}@{}}
\caption{Variables used in this study.}
\label{Variables table}\\
\toprule
Variables                     & Description                                                                                                                                                                                              \\* \midrule
\endfirsthead
\multicolumn{2}{c}%
{{\bfseries Table \thetable\ continued from previous page}} \\
\toprule
Variables                     & Description                                                                                                                                                                                              \\* \midrule
\endhead
\bottomrule
\endfoot
\endlastfoot
loan\_amnt                  & Amount of the loan applied for by the borrower.                                                                                                                                               \\
acc\_now\_delinq                        & The number of accounts on which the borrower is now delinquent.                                                                                                                     \\
int\_rate                   & Interest Rate on the loan                                                                                                                                                                                \\
installment                 & The monthly payment owed by the borrower if the loan originates.                                                                                                                                         \\
annual\_inc                 & The self-reported annual income provided by the borrower during registration.                                                                                                                            \\
emp\_length                 & employment length in years.                                                                                                                                                                             \\
verification\_status        & Indicates if income was verified by LC~\footnote{Lending Club}, not verified, or if the income source was verified                                                                                                               \\
dti                         & The borrower’s total monthly debt payments on the total debt obligations, excluding mortgage and the requested LC loan, divided by the borrower’s self-reported monthly income. \\
delinq\_2yrs                & The number of 30+ days past-due incidences of delinquency in the borrower's credit file for the past 2 years                                                                                             \\
average\_fico            & The average value of the upper and lower boundary range of the borrower’s FICO at loan origination.                                                                                                                            \\
inq\_last\_6mths            & The number of inquiries in past 6 months (excluding auto and mortgage inquiries).                                                                                                                         \\
open\_acc                   & The number of open credit lines in the borrower's credit file.                                                                                                                                            \\
pub\_rec                    & Number of derogatory public records.                                                                                                                                                                      \\
revol\_util                 & Revolving line utilization rate, or the amount of credit the borrower is using relative to all available revolving credit.                                                                               \\
total\_acc                  & The total number of credit lines currently in the borrower's credit file.                                                                                                                                 \\
chargeoff\_within\_12\_mths & Number of charge-offs within 12 months.                                                                                                                                                                   \\
delinq\_amnt                & The past-due amount owed for the accounts on which the borrower is now delinquent                                                                                                                       \\
mort\_acc                   & Number of mortgage accounts                                                                                                                                                                             \\
pub\_rec\_bankruptcies      & Number of public record bankruptcies                                                                                                                                                                     \\
tax\_liens                  & Number of tax liens                                                                                                                                                                                     \\
grade                       & LC assigned loan grade. The values are: [A,B,C,D,E,F,G]                                                                                                                                                                                  \\
subgrade                    & LC assigned loan subgrade. The values are: [A1,\dots,A5,\dots,G1,\dots,G5]                                                                                                                                                                               \\
home\_ownership             & The home ownership status provided by the borrower during registration or obtained from the credit report. The values are: RENT, OWN, MORTGAGE                                                    \\
purpose                     & A category provided by the borrower for the loan request such as: Debt consolidation, small business, vacation, etc.                                                                                                                                                 \\
initial\_list\_status       & The initial listing status of the loan. The values are: Whole, Fractional                                                                                                                                       \\
loan\_status                & Current status of the loan (Target variable)                                                                                                                                                                               \\* \bottomrule
\end{longtable}

\subsection{Evaluation measures}
To evaluate the performance of the techniques studied in this paper, we consider six performance measures: Accuracy, Area under the ROC curve (AUC), H-measure, F-measure, Brier score, and G-mean. The performance measures we have selected are often used in credit scoring, and they can cover almost all important aspects of classification models performance. The accuracy evaluates the correctness of categorical predictions, the AUC and H-measure assess discriminatory ability, and the Brier score assesses the accuracy of probability predictions. The G-mean evaluates the balance between the classification results of positive and negative classes. In the following, we explain how to compute these metrics from the confusion matrix (see Table~\ref{confusion matrix})

\begin{table}[h]
\centering
\caption{Confusion matrix}
\label{confusion matrix}
\begin{tabular}{llll} 
\toprule
     &          & \multicolumn{2}{c}{Predicted}              \\ 
\cline{3-4}
     &          & Positive            & Negative             \\ 
\hline
Real & Positive & True Positive (TP)  & False Negative (FN)  \\
     & Negative & False Positive (FP) & True Negative (TN)   \\
\bottomrule
\end{tabular}
\end{table}

\subsubsection{Accuracy}
Accuracy is the correct prediction sample size divided by the total testing sample size, as shown in equation \ref{acc}.

\begin{flalign}\label{acc}
\text{Accuracy} = \frac{TP + TN}{TP + FN + FP + TN}
\end{flalign}

\subsubsection{F-measure}
F-measure is the harmonic mean of precision and  recall where precision and recall measures are computed via equation \ref{precision} and \ref{recall} respectively. 

\begin{flalign}\label{fmeasure}
\text{F-measure} = \frac{2 \times Precision \times Recall}{Precision + Recall}
\end{flalign}

\begin{flalign}\label{precision}
\text{Precision} = \frac{TP}{TP + FP}
\end{flalign}

\begin{flalign}\label{recall}
\text{Recall} = \text{Sensitivity} = \text{TPR} = \frac{TP}{TP + FN}
\end{flalign}

\subsubsection{G-mean}
Geometric mean is a comprehensive evaluation method based on sensitivity and specificity presented in equation \ref{gmean} \citep{kubat1997addressing}. 
Sensitivity shows the percentage of true positives, similar to precision, while specificity computes the percentage of true negative. Sensitivity and specificity are calculated as shown in equation \ref{recall} and \ref{spec} respectively. 
A higher value of G-mean shows the balance between classes is reasonable and classification model has good performance in binary classification. 

\begin{flalign}\label{gmean}
\text{G-mean} = \sqrt{Sensitivity \times Specificity}
\end{flalign}

\begin{flalign}\label{spec}
\text{Specificity} = \frac{TN}{TN + FP}
\end{flalign}

\subsubsection{AUC}
The AUC is obtained from the Receiver Operating Characteristic (ROC) curve representing the area under the ROC curve \citep{fawcett2004roc}. The horizontal axis of ROC curve represents the false-positive rate (1-specificity) and the vertical axis of this case represents the true-positive ate (sensitivity). In this curve the true positive rate is plotted against the false positive rate at various cut-off values where each point on the ROC curve signifies a pair of TPR and FPR that corresponds to a certain threshold. The AUC of a classifier equals the probability that a randomly chosen positive case receives a score higher than a randomly chosen negative case, and its value is in the range of [0,1].

\subsubsection{H-measure}
The H-measure metric was proposed by \citet{hand2009measuring}. H-measure uses a pre-specified beta distribution to represent the misclassification cost distribution function.

\subsubsection{Brier score}
The Brier score measures the accuracy of probabilistic prediction. The Brier score ranges from 0 (perfect probabilistic prediction) to 1 (poor prediction) as the following equation. Where N is the number of samples, $s_i$ is the probability prediction of classification and $y_i$ is the true label of the sample.

\begin{flalign}\label{bs}
\text{Brier score} = \frac{1}{N} \sum_{i=1}^N (s_i - y_i)^2
\end{flalign}

\subsection{Experiments}
\subsubsection{DS techniques evaluation}\label{experiment1}
Dynamic selection techniques select the most competent classifiers for each query sample in its region of competence. Based on this definition, to create the pool of classifiers, we used bootstrap aggregation (Bagging) using k nearest neighbors, support vector machine, Gaussian naive Bayes, and multilayer perceptron as the base classifiers. We also implemented DS on random forest due to its high performance in credit scoring classification~\citep{lessmann2015benchmarking}. In addition to homogeneous classifier pools, we also created a heterogeneous pool using all previously mentioned base classifiers as well as random forest. The number of base classifiers in the pool as well as the number of trees in the random forest, were determined by a grid search (values between 10 to 200). We held aside twenty-five percent of the training set for the DSEL. The region of competence for each test sample was calculated based on the samples in the DSEL. We ran all the experiments using scikit-learn library in python for training the base classifiers and the pool. We used the DESlib library in python by \citet{cruz2018deslib} to implement the DS techniques on the created pool of classifiers. Hyper-parameters of the base classifiers and the DS techniques were set to the default values.

\subsubsection{DS performance on imbalanced data}\label{experiment2}
The imbalance ratio of a data set is defined as the number of samples in the majority class divided by the number of minority class samples. Credit scoring data sets often have a high imbalance ratio, which makes classification a challenging task. \citet{cruz2018dynamic} state that since dynamic selection techniques perform locally, the classifiers that are selected to classify each query sample are not affected by all samples in the data set and only take into account the neighborhood of the query sample. This notion suggests that dynamic selection techniques might be robust against imbalanced data sets which we will validate by the following experiments.

To evaluate the robustness of dynamic selection techniques in classifying imbalanced data sets, we modify the imbalance ratio of our data set by under-sampling majority class. The imbalance ratio for the original training set is $5.8$. We under-sample majority class training samples to create data sets with varying degrees of imbalance from a completely balanced data set (i.e., imbalance ratio=1) to the original data set. The imbalance ratio and the number of samples in each data set are shown in Table~\ref{IR}. 
For each modified data set, the same structure for pools of classifiers was considered, thus all the experiments on the original data set were repeated on modified data sets. The number of classifiers in the pools was set as the optimized number of classifiers for the original data set with default values of hyper-parameters for each base classifier.     

\begin{table}[H]
\centering
\caption{The imbalance ratio and the number of samples in modified data sets created by under-sampling}
\label{IR}
\begin{tabular}{lcc} 
\toprule
Imbalance Ratio         & \multicolumn{1}{l}{No. majority samples} & \multicolumn{1}{l}{No. minority samples}  \\ 
\hline
1                       & 23863                                    & 23863                                     \\
2                       & 47726                                    & 23863                                     \\
3                       & 71589                                    & 23863                                     \\
4                       & 95452                                    & 23863                                     \\
5                       & 119315                                   & 23863                                     \\
5.8 (Original data set) & 138372                                   & 23863                                     \\
\bottomrule
\end{tabular}
\end{table}

\section{Results and Discussion}
The results of our experiments in section \ref{experiment1} and \ref{experiment2} are represented in Tables 12-15 (the best three DS techniques are in bold across all evaluation measures). Based on the obtained results we discuss our findings regarding the following aspects:
\subsection{On the impact of DS}
The hypothesis behind employing dynamic selection techniques for to ensemble models is to increase their classification ability by choosing the most competent classifiers for each test sample. From the results of our experiments on the real-life credit data of Lending Club customers, it is observed that DS techniques are able to improve the pool of classifiers’ performance. Table \ref{Table DS discussion} shows the average performance of the top 3 DS techniques regarding different performance measures compared to the performance of the pool in the original data set. The obtained results demonstrate that DS techniques can boost the performance of ensembles. The improvements are particularly observed in G-mean and F1 score, which represents the classification ability of classifiers in the presence of cost-sensitive and imbalanced training sets. This result indicates that we can validate the statement of \citet{cruz2018dynamic} that dynamic selection techniques are great candidates for classifying imbalanced data sets specifically in the context of highly imbalanced credit scoring problems.

\begin{table}[H]
\centering
\caption{The average performance of top 3 DS techniques compared to the pool of classifiers.}
\label{Table DS discussion}
\resizebox{\linewidth}{!}{%
\begin{tabular}{llccllcc} 
\toprule
Classifier & \multicolumn{1}{c}{\begin{tabular}[c]{@{}c@{}}Evaluation \\measure \end{tabular}} & \begin{tabular}[c]{@{}c@{}}Classifiers' \\pool \end{tabular} & \begin{tabular}[c]{@{}c@{}}Average top 3 \\DS techniques \end{tabular} & Classifier & \multicolumn{1}{c}{\begin{tabular}[c]{@{}c@{}}Evaluation \\measure \end{tabular}} & \begin{tabular}[c]{@{}c@{}}Classifiers' \\pool \end{tabular} & \begin{tabular}[c]{@{}c@{}}Average top 3 \\DS techniques \end{tabular} \\ 
\hline
\multirow{6}{*}{\begin{tabular}[c]{@{}l@{}}GNB \\~\\~\\~\\~\\~\\~\\ \end{tabular}} & Acc & 0.7728 & \textbf{0.8212 }  & \multirow{6}{*}{\begin{tabular}[c]{@{}l@{}}Heterogeneous\\pool \\~\\~\\~\\~\\~\\ \end{tabular}} & Acc & 0.8459 & \textbf{0.8510 }  \\
 & AUC & 0.6879 & 0.6763 &  & AUC & 0.6813 & 0.6794 \\
 & F1 & 0.3171 & 0.2876 &  & F1 & 0.1112 & \textbf{0.2144 }  \\
 & G-mean & 0.5483 & 0.5083 &  & G-mean & 0.2527 & \textbf{0.3967 }  \\
 & H\_measure & 0.1105 & 0.1019 &  & H\_measure & 0.1065 & 0.1030 \\
 & Brier score & 0.1816 & \textbf{0.1656 }  &  & Brier score & 0.1231 & \textbf{0.1221 }  \\
\multirow{6}{*}{\begin{tabular}[c]{@{}l@{}}RF \\~\\~\\~\\~\\~\\~\\ \end{tabular}} & Acc & 0.8519 & \textbf{0.8519 }  & \multirow{6}{*}{\begin{tabular}[c]{@{}l@{}}MLP \\~\\~\\~\\~\\~\\~\\ \end{tabular}} & Acc & 0.8494 & \textbf{ 0.8499}  \\
 & AUC & 0.6829 & \textbf{0.6837 }  &  & AUC & 0.6837 & 0.6824 \\
 & F1 & 0.0369 & \textbf{0.2144 }  &  & F1 & 0.0656 & \textbf{ 0.1804}  \\
 & G-mean & 0.1382 & \textbf{0.4360 }  &  & G-mean & 0.1879 & \textbf{ 0.3532}  \\
 & H\_measure & 0.1071 & \textbf{0.1084 }  &  & H\_measure & 0.1062 & 0.1047 \\
 & Brier score & 0.1198 & 0.1204 &  & Brier score & 0.1204 & 0.1205 \\
\bottomrule
\end{tabular}
}
\end{table}

\subsection{On the impact of DS on imbalanced data sets}
The results of our study shows that generally the accuracy increases with the increase in the imbalance ratio of the training data sets. Simultaneously, the G-mean measure, which reflects the ability of a classifier to distinguish between majority and minority classes, decreases in the results. This concurrent might indicate that as the classification accuracy increases, in higher imbalanced data sets, the ability of distinguishing between negative (minority) class and positive (majority) class of the data set decreases. We can argue that the increase in the accuracy measure is a consequence of encountering more negative samples which increases the ability of the algorithms to classify them correctly. 

Applying dynamic selection techniques on the pool of classifiers across different versions of the data set, can increase the ability of the ensemble to distinguish between minority and majority classes. The increase of the F1 measure, G-mean and AUC metric along with improvement of accuracy can indicate that DS techniques are able to improve the performance of the pool, especially in data sets with higher imbalance ratios. Tables \ref{DS IR1 discussion} to \ref{DS IR5 discussion} show the average results of the top 3 DS techniques compared to the pool of classifiers’ performance regarding different imbalance ratios.
\begin{table}[H]
\centering
\caption{The average performance of top 3 DS techniques compared to the pool of classifiers in data set (Imbalance Ratio=1).}
\label{DS IR1 discussion}
\resizebox{\linewidth}{!}{%
\begin{tabular}{llccllcc} 
\toprule
Classifier & \begin{tabular}[c]{@{}l@{}}Evaluation\\\textasciitilde{}measure \end{tabular} & \multicolumn{1}{l}{\begin{tabular}[c]{@{}l@{}}Classifiers\\\textasciitilde{}pool \end{tabular}} & \multicolumn{1}{l}{\begin{tabular}[c]{@{}l@{}}Average top 3\\\textasciitilde{}DS techniques \end{tabular}} & Classifier & \begin{tabular}[c]{@{}l@{}}Evaluation\\\textasciitilde{}measure \end{tabular} & \multicolumn{1}{l}{\begin{tabular}[c]{@{}l@{}}Classifiers\\pool \end{tabular}} & \multicolumn{1}{l}{\begin{tabular}[c]{@{}l@{}}Average top 3\\DS techniques \end{tabular}} \\ 
\hline
\multirow{6}{*}{\begin{tabular}[c]{@{}l@{}}GNB\\~\\~\\~\\~\\~\\~\\ \end{tabular}} & Acc & 0.8251 & \textbf{0.8287}  & \multirow{6}{*}{\begin{tabular}[c]{@{}l@{}}Heterogeneous\\pool\\~\\~\\~\\~\\~\\ \end{tabular}} & Acc & 0.7406 & \textbf{0.7443}  \\
 & AUC & 0.6854 & \textbf{0.6870}  &  & AUC & 0.6856 & 0.6827 \\
 & F1 & 0.2298 & \textbf{0.2910}  &  & F1 & 0.3310 & \textbf{0.3464}  \\
 & G-mean & 0.4062 & \textbf{0.4978}  &  & G-mean & 0.5861 & \textbf{0.6373}  \\
 & H\_measure & 0.1072 & \textbf{0.1098}  &  & H\_measure & 0.1094 & 0.1071 \\
 & Brier score & 0.1587 & 0.1658 &  & Brier score & 0.1813 & 0.1856 \\
\multirow{6}{*}{\begin{tabular}[c]{@{}l@{}}RF \\~\\~\\~\\~\\~\\~\\ \end{tabular}} & Acc & 0.6233 & \textbf{0.6265}  & \multirow{6}{*}{\begin{tabular}[c]{@{}l@{}}MLP \\~\\~\\~\\~\\~\\~\\ \end{tabular}} & Acc & 0.6225 & 0.6283 \\
 & AUC & 0.6934 & 0.6910 &  & AUC & 0.6758 & 0.6776 \\
 & F1 & 0.3435 & 0.3425 &  & F1 & 0.3346 & 0.3338 \\
 & G-mean & 0.6397 & 0.6387 &  & G-mean & 0.6296 & 0.6284 \\
 & H\_measure & 0.1151 & 0.1131 &  & H\_measure & 0.0965 & 0.0988 \\
 & Brier score & 0.2227 & 0.2250 &  & Brier score & 0.2303 & 0.2312 \\
\bottomrule
\end{tabular}
}
\end{table}

\begin{table}[H]
\centering
\caption{The average performance of top 3 DS techniques compared to the pool of classifiers in data set (Imbalance Ratio=2).}
\label{DS IR2 discussion}
\resizebox{\linewidth}{!}{%
\begin{tabular}{llccllcc} 
\toprule
Classifier & \begin{tabular}[c]{@{}l@{}}Evaluation\\measure \end{tabular} & \multicolumn{1}{l}{\begin{tabular}[c]{@{}l@{}}Classifiers\\pool \end{tabular}} & \multicolumn{1}{l}{\begin{tabular}[c]{@{}l@{}}Average top 3\\DS techniques \end{tabular}} & Classifier & \begin{tabular}[c]{@{}l@{}}Evaluation\\measure \end{tabular} & \multicolumn{1}{l}{\begin{tabular}[c]{@{}l@{}}Classifiers\\pool \end{tabular}} & \multicolumn{1}{l}{\begin{tabular}[c]{@{}l@{}}Average top 3\\DS techniques \end{tabular}} \\ 
\hline
\multirow{6}{*}{\begin{tabular}[c]{@{}l@{}}GNB\\~\\~\\~\\~\\~\\~\\ \end{tabular}} & Acc & 0.7996 & 0.7954 & \multirow{6}{*}{\begin{tabular}[c]{@{}l@{}}Heterogeneous\\pool \\~\\~\\~\\~\\ \end{tabular}} & Acc & 0.8354 & \textbf{0.8487}  \\
 & AUC & 0.6872 & 0.6824 &  & AUC & 0.6820 & 0.6814 \\
 & F1 & 0.2797 & \textbf{0.3273}  &  & F1 & 0.1944 & \textbf{0.2391}  \\
 & G-mean & 0.4839 & \textbf{0.5908}  &  & G-mean & 0.3580 & \textbf{0.4347}  \\
 & H\_measure & 0.1097 & 0.1050 &  & H\_measure & 0.1072 & 0.1062 \\
 & Brier score & 0.1641 & 0.1760 &  & Brier score & 0.1273 & \textbf{0.1231}  \\
\multirow{6}{*}{\begin{tabular}[c]{@{}l@{}}RF \\~\\~\\~\\~\\~\\~\\ \end{tabular}} & Acc & 0.8017 & 0.8006 & \multirow{6}{*}{\begin{tabular}[c]{@{}l@{}}MLP \\~\\~\\~\\~\\~\\~\\ \end{tabular}} & Acc & 0.7845 & \textbf{ 0.7859}  \\
 & AUC & 0.6908 & 0.6893 &  & AUC & 0.6812 & 0.6800 \\
 & F1 & 0.3028 & 0.3016 &  & F1 & 0.2974 & 0.2950 \\
 & G-mean & 0.5085 & \textbf{0.5364}  &  & G-mean & 0.5164 & \textbf{ 0.5392}  \\
 & H\_measure & 0.1138 & 0.1128 &  & H\_measure & 0.1032 & 0.1025 \\
 & Brier score & 0.1535 & 0.1446 &  & Brier score & 0.1561 & \textbf{ 0.1530}  \\
\bottomrule
\end{tabular}
}
\end{table}

\begin{table}[H]
\centering
\caption{The  average  performance  of  top  3  DS  techniques  compared  to  the pool of classifiers in data set (Imbalance Ratio=3).}
\label{DS IR3 discussion}
\resizebox{\linewidth}{!}{%
\begin{tabular}{llccllcc} 
\toprule
Classifier & \begin{tabular}[c]{@{}l@{}}Evaluation\\measure \end{tabular} & \multicolumn{1}{l}{\begin{tabular}[c]{@{}l@{}}Classifiers\\pool \end{tabular}} & \multicolumn{1}{l}{\begin{tabular}[c]{@{}l@{}}Average top 3\\DS techniques \end{tabular}} & Classifier & \begin{tabular}[c]{@{}l@{}}Evaluation\\measure \end{tabular} & \multicolumn{1}{l}{\begin{tabular}[c]{@{}l@{}}Classifiers\\pool \end{tabular}} & \multicolumn{1}{l}{\begin{tabular}[c]{@{}l@{}}Average top 3\\DS techniques \end{tabular}} \\ 
\hline
\multirow{6}{*}{\begin{tabular}[c]{@{}l@{}}GNB \\~\\~\\~\\~\\~\\~\\ \end{tabular}} & Acc & 0.7956 & \textbf{0.8121}  & \multirow{6}{*}{\begin{tabular}[c]{@{}l@{}}Heterogeneous\\pool \\~\\~\\~\\~\\ \end{tabular}} & Acc & 0.8138 & \textbf{0.8406}  \\
 & AUC & 0.6870 & 0.6804 &  & AUC & 0.6837 & 0.6824 \\
 & F1 & 0.2833 & \textbf{0.3263}  &  & F1 & 0.2762 & 0.2751 \\
 & G-mean & 0.4914 & \textbf{0.5858}  &  & G-mean & 0.4678 & \textbf{0.4897}  \\
 & H\_measure & 0.1091 & 0.1037 &  & H\_measure & 0.1085 & 0.1062 \\
 & Brier score & 0.1720 & \textbf{0.1687}  &  & Brier score & 0.1364 & \textbf{0.1305}  \\
\multirow{6}{*}{\begin{tabular}[c]{@{}l@{}}RF \\~\\~\\~\\~\\~\\~\\ \end{tabular}} & Acc & 0.8370 & \textbf{0.8389}  & \multirow{6}{*}{\begin{tabular}[c]{@{}l@{}}MLP \\~\\~\\~\\~\\~\\~\\ \end{tabular}} & Acc & 0.8293 & \textbf{ 0.8307}  \\
 & AUC & 0.6892 & 0.6887 &  & AUC & 0.6796 & 0.6786 \\
 & F1 & 0.1785 & \textbf{0.2500}  &  & F1 & 0.2087 & \textbf{ 0.2532}  \\
 & G-mean & 0.3389 & \textbf{0.5072}  &  & G-mean & 0.3791 & \textbf{ 0.4805}  \\
 & H\_measure & 0.1119 & 0.1117 &  & H\_measure & 0.1013 & 0.1003 \\
 & Brier score & 0.1325 & \textbf{0.1283}  &  & Brier score & 0.1335 & \textbf{ 0.1321}  \\
\bottomrule
\end{tabular}
}
\end{table}

\begin{table}[H]
\centering
\caption{The average performance of top 3 DS techniques compared to the pool of classifiers in data set (Imbalance Ratio=4).}
\label{DS IR4 discussion}
\resizebox{\linewidth}{!}{%
\begin{tabular}{llccllcc} 
\toprule
Classifier & \begin{tabular}[c]{@{}l@{}}Evaluation\\measure \end{tabular} & \multicolumn{1}{l}{\begin{tabular}[c]{@{}l@{}}Classifiers\\pool \end{tabular}} & \multicolumn{1}{l}{\begin{tabular}[c]{@{}l@{}}Average top 3\\DS techniques \end{tabular}} & Classifier & \begin{tabular}[c]{@{}l@{}}Evaluation\\measure \end{tabular} & \multicolumn{1}{l}{\begin{tabular}[c]{@{}l@{}}Classifiers\\pool \end{tabular}} & \multicolumn{1}{l}{\begin{tabular}[c]{@{}l@{}}Average top 3\\DS techniques \end{tabular}} \\ 
\hline
\multirow{6}{*}{\begin{tabular}[c]{@{}l@{}}GNB\\~\\~\\~\\~\\~\\ \end{tabular}} & Acc & 0.7944 & \textbf{0.8117}  & \multirow{6}{*}{\begin{tabular}[c]{@{}l@{}}Heterogeneous\\pool \\~\\~\\~\\~\\ \end{tabular}} & Acc & 0.8138 & \textbf{0.8406}  \\
 & AUC & 0.6874 & 0.6762 &  & AUC & 0.6837 & 0.6824 \\
 & F1 & 0.2847 & \textbf{0.3266}  &  & F1 & 0.2762 & 0.2751 \\
 & G-mean & 0.4939 & \textbf{0.5848}  &  & G-mean & 0.4678 & \textbf{0.4897}  \\
 & H\_measure & 0.1097 & 0.1015 &  & H\_measure & 0.1085 & 0.1062 \\
 & Brier score & 0.1745 & \textbf{0.1707}  &  & Brier score & 0.1364 & \textbf{0.1305}  \\
\multirow{6}{*}{\begin{tabular}[c]{@{}l@{}}RF \\~\\~\\~\\~\\~\\~\\ \end{tabular}} & Acc & 0.8485 & \textbf{0.8490}  & \multirow{6}{*}{\begin{tabular}[c]{@{}l@{}}MLP \\~\\~\\~\\~\\~\\~\\ \end{tabular}} & Acc & 0.8443 & \textbf{ 0.8453}  \\
 & AUC & 0.6867 & \textbf{0.6871}  &  & AUC & 0.6835 & 0.6825 \\
 & F1 & 0.0992 & \textbf{0.2332}  &  & F1 & 0.1233 & \textbf{ 0.2226}  \\
 & G-mean & 0.2355 & \textbf{0.4747}  &  & G-mean & 0.2687 & \textbf{ 0.4214}  \\
 & H\_measure & 0.1101 & \textbf{0.1109}  &  & H\_measure & 0.1058 & 0.1044 \\
 & Brier score & 0.1242 & \textbf{0.1240}  &  & Brier score & 0.1238 & \textbf{ 0.1236}  \\
\bottomrule
\end{tabular}
}
\end{table}

\begin{table}[H]
\centering
\caption{The average performance of top 3 DS techniques compared to the poolof classifiers in data set (Imbalance Ratio=5).}
\label{DS IR5 discussion}
\resizebox{\linewidth}{!}{%
\begin{tabular}{llccllcc} 
\toprule
Classifier & \begin{tabular}[c]{@{}l@{}}Evaluation\\measure \end{tabular} & \multicolumn{1}{l}{\begin{tabular}[c]{@{}l@{}}Classifiers\\pool \end{tabular}} & \multicolumn{1}{l}{\begin{tabular}[c]{@{}l@{}}Average top 3\\DS techniques \end{tabular}} & Classifier & \begin{tabular}[c]{@{}l@{}}Evaluation\\measure \end{tabular} & \multicolumn{1}{l}{\begin{tabular}[c]{@{}l@{}}Classifiers\\pool \end{tabular}} & \multicolumn{1}{l}{\begin{tabular}[c]{@{}l@{}}Average top 3\\DS techniques \end{tabular}} \\ 
\hline
\multirow{6}{*}{\begin{tabular}[c]{@{}l@{}}GNB\\~\\~\\~\\~\\~\\~\\ \end{tabular}} & Acc & 0.7121 & \textbf{0.8105}  & \multirow{6}{*}{\begin{tabular}[c]{@{}l@{}}Heterogeneous\\pool\\~\\~\\~\\~\\ \end{tabular}} & Acc & 0.8438 & \textbf{0.8503}  \\
 & AUC & 0.6876 & 0.6841 &  & AUC & 0.6820 & 0.6796 \\
 & F1 & 0.3370 & 0.3184 &  & F1 & 0.1423 & \textbf{0.2260}  \\
 & G-mean & 0.6082 & 0.5641 &  & G-mean & 0.2919 & \textbf{0.4148}  \\
 & H\_measure & 0.1102 & 0.1072 &  & H\_measure & 0.1073 & 0.1031 \\
 & Brier score & 0.2218 & \textbf{0.1754}  &  & Brier score & 0.1245 & \textbf{0.1223}  \\
\multirow{6}{*}{\begin{tabular}[c]{@{}l@{}}RF \\~\\~\\~\\~\\~\\~\\ \end{tabular}} & Acc & 0.8508 & \textbf{0.8509}  & \multirow{6}{*}{\begin{tabular}[c]{@{}l@{}}MLP \\~\\~\\~\\~\\~\\~\\ \end{tabular}} & Acc & 0.8484 & \textbf{ 0.8487}  \\
 & AUC & 0.6851 & \textbf{0.6855}  &  & AUC & 0.6849 & 0.6836 \\
 & F1 & 0.0462 & \textbf{0.2228}  &  & F1 & 0.0891 & \textbf{ 0.1943}  \\
 & G-mean & 0.1556 & \textbf{0.4520}  &  & G-mean & 0.2222 & \textbf{ 0.3755}  \\
 & H\_measure & 0.1085 & \textbf{0.1094}  &  & H\_measure & 0.1075 & 0.1060 \\
 & Brier score & 0.1209 & 0.1216 &  & Brier score & 0.1210 & \textbf{ 0.1210}  \\
\bottomrule
\end{tabular}
}
\end{table}

\subsection{On comparison of DS techniques}
The F1 measure considers the trade-off between precision and recall measures (i.e., fewer false positives and fewer false negatives). We use the F1 measure to compare the results of different dynamic selection techniques on the original data set. In our comparison of DS techniques, we have ranked their performance based on the F1 measure as shown in Table \ref{Ranks table}. From the rank of DS techniques in the ensemble pool of each classifier in the original data set, it is observed that the overall top 3 classification techniques are \textit{GNB\_bag}, \textit{GNB\_posteriori}, and \textit{Pool\_posteriori}. It is interesting to mention that more complex DS techniques such as Meta-DES do not seem to perform as efficiently as less complex DS techniques such as A Posteriori. This may be due to the fact that the Meta-DES technique is considered to be most effective in small sample size classification problems \citep{cruz2015meta}, whereas our data set contains more than 160,000 training samples with high dimensionality.  
\begin{table}[H]
\centering
\caption{Ranks of different techniques based on the F measure in the original data set.}
\label{Ranks table}
\resizebox{\linewidth}{!}{%
\begin{tabular}{lcclcc} 
\toprule
Classifier & F1 measure\textasciitilde{} & Rank & Classifier & F1 measure\textasciitilde{} & Rank \\ 
\hline
GNB\_bag & 0.3171 & 1 & Pool & 0.1112 & 34 \\
GNB\_posteriori & 0.2989 & 2 & Pool\_posteriori & 0.2932 & 3 \\
GNB\_priori & 0.2379 & 14 & Pool\_priori & 0.1035 & 35 \\
GNB\_lca & 0.2727 & 8 & Pool\_lca & 0.0493 & 51 \\
GNB\_mcb & 0.2496 & 12 & Pool\_mcb & 0.1200 & 32 \\
GNB\_mla & 0.2727 & 8 & Pool\_mla & 0.0497 & 50 \\
GNB\_ola & 0.2467 & 13 & Pool\_ola & 0.1246 & 31 \\
GNB\_rank & 0.2738 & 7 & Pool\_rank & 0.1825 & 22 \\
GNB\_metades & 0.2264 & 16 & Pool\_metades & 0.0602 & 45 \\
GNB\_descluster & 0.2701 & 11 & Pool\_descluster & 0.0105 & 60 \\
GNB\_desknn & 0.2765 & 6 & Pool\_desknn & 0.0356 & 58 \\
GNB\_desp & 0.2326 & 15 & Pool\_desp & 0.0800 & 36 \\
GNB\_knop & 0.2809 & 5 & Pool\_knop & 0.0533 & 49 \\
GNB\_kne & 0.2720 & 10 & Pool\_kne & 0.1674 & 25 \\
GNB\_knu & 0.2829 & 4 & Pool\_knu & 0.0558 & 47 \\
RF\_indv & 0.0369 & 57 & MLP\_bag & 0.0656 & 40 \\
RF\_posteriori & 0.1469 & 29 & MLP\_posteriori & 0.0478 & 52 \\
RF\_priori & 0.2177 & 17 & MLP\_priori & 0.1135 & 33 \\
RF\_lca & 0.1494 & 27 & MLP\_lca & 0.0657 & 38 \\
RF\_mcb & 0.2129 & 18 & MLP\_mcb & 0.1850 & 21 \\
RF\_mla & 0.1494 & 27 & MLP\_mla & 0.0657 & 38 \\
RF\_ola & 0.2116 & 20 & MLP\_ola & 0.1756 & 24 \\
RF\_rank & 0.2127 & 19 & MLP\_rank & 0.1808 & 23 \\
RF\_metades & 0.0370 & 56 & MLP\_metades & 0.0642 & 41 \\
RF\_descluster & 0.0454 & 53 & MLP\_descluster & 0.0630 & 43 \\
RF\_desknn & 0.0540 & 48 & MLP\_desknn & 0.0620 & 44 \\
RF\_desp & 0.0411 & 54 & MLP\_desp & 0.0694 & 37 \\
RF\_knop & 0.0216 & 59 & MLP\_knop & 0.0596 & 46 \\
RF\_kne & 0.1496 & 26 & MLP\_kne & 0.1438 & 30 \\
RF\_knu & 0.0384 & 55 & MLP\_knu & 0.0631 & 42 \\
\bottomrule
\end{tabular}
}
\end{table}
\section{Conclusion}
In this paper, we set out to find out the effectiveness of dynamic selection techniques to improve the classification ability of both homogeneous and heterogeneous ensemble models in credit scoring. Dynamic selection techniques, are applied on a pool of classifiers and select the most competent classifiers for each test query individually based on a pre-specified competency measure. 

We have trained 14 different dynamic selection techniques trained on a real-life credit scoring data set. Also, to inspect the robustness of DS techniques in imbalanced environments, we have formed different training sets from the original training set by under-sampling the majority class samples. DS techniques were then trained on each of the data sets and used to classify the test set. 

Based on the results of our experiments, we can conclude that dynamic selection techniques are able to boost the performance of ensemble models. DS techniques are mostly effective in increasing the G-mean and F1 measure which determine the classifiers' ability to perform well in learning from imbalanced data sets. Our experiments validate \citet{cruz2018dynamic} assumption about the robustness of dynamic selection techniques in dealing with imbalanced data sets. 

Based on the results of the F1 measure, we can conclude that less complex dynamic selection techniques such as A Posteriori, Knora Union and Eliminate  are able to perform better as opposed to complicated techniques such as Meta-DES and DES Performance. Which may be due to the fact that large and high dimensional data sets are used in credit scoring.

The majority of DS techniques use k-NN algorithm to obtain the region of competence for each test sample, therefore dynamic selection techniques have higher complexity compared to the pool of classifiers. One of the limitation of our study is that due to computation power, we have used the default value of the hyper-parameters both in our classifiers and DS techniques due to our computation limitations. We believe that by hyper-parameter optimization the results of dynamic selection techniques may improve classification more significantly. Additionally, due to the limitaitons of our computation power, we were not able to train the pool of support vector machine and k-nearest neighbors on our real-life data set. 

The high complexity of DS techniques may be one of the drawbacks of employing these techniques especially in high dimensional data sets. We believe our experiments on a large and high dimensional data set can be used as the starting point of other studies both in the context of credit scoring and other classification tasks with large data sets.

\begin{sidewaystable}
\centering
\caption{GNB results (Number of base classifiers=80)}
\label{gnb_results}
\resizebox{\linewidth}{!}{%
\begin{tabular}{clcccccccccccccccc} 
\toprule
\multirow{2}{*}{Imbalance Ratio} & \multicolumn{1}{c}{\multirow{2}{*}{Evaluation measure}} & \multicolumn{16}{c}{Classification technique} \\ 
\cline{3-18}
 & \multicolumn{1}{c}{} & GNB\_indv & GNB\_bag & GNB\_posteriori & GNB\_priori & GNB\_lca & GNB\_mcb & GNB\_mla & GNB\_ola & GNB\_rank & GNB\_metades & GNB\_descluster & GNB\_desknn & GNB\_desp & GNB\_knop & GNB\_kne & GNB\_knu \\ 
\hline
\multirow{6}{*}{\begin{tabular}[c]{@{}c@{}}1\\~\\~\\~\\~\\~\\~\\ \end{tabular}} & Acc & 0.8217 & 0.8251 & \textbf{0.8244 }  & 0.7981 & \textbf{0.8309 }  & 0.7990 & \textbf{0.8309 }  & 0.7983 & 0.8033 & 0.8029 & 0.8194 & 0.8222 & 0.8000 & 0.8215 & 0.8039 & 0.8223 \\
 & AUC & 0.6866 & 0.6854 & 0.6732 & 0.6834 & 0.6601 & 0.6852 & 0.6601 & 0.6831 & 0.6765 & 0.6848 & \textbf{0.6873 }  & \textbf{0.6871 }  & \textbf{0.6868 }  & 0.6863 & 0.6858 & 0.6864 \\
 & F1 & 0.2369 & 0.2298 & 0.2258 & \textbf{0.2888 }  & 0.1982 & \textbf{0.2922 }  & 0.1982 & \textbf{0.2919 }  & 0.2777 & 0.2703 & 0.2524 & 0.2461 & 0.2868 & 0.2438 & 0.2760 & 0.2459 \\
 & G-mean & 0.4170 & 0.4062 & 0.4023 & \textbf{0.4956 }  & 0.3661 & \textbf{0.4988 }  & 0.3661 & \textbf{0.4990 }  & 0.4786 & 0.4705 & 0.4363 & 0.4268 & 0.4917 & 0.4248 & 0.4762 & 0.4265 \\
 & H\_measure & 0.1073 & 0.1072 & 0.0950 & 0.1051 & 0.0860 & 0.1073 & 0.0860 & 0.1054 & 0.0996 & 0.1063 & \textbf{0.1103 }  & \textbf{0.1096 }  & \textbf{0.1095 }  & 0.1084 & 0.1071 & 0.1089 \\
 & Brier score & 0.1754 & 0.1587 & 0.1741 & 0.1940 & 0.1675 & 0.1893 & 0.1675 & 0.1918 & 0.1872 & 0.1818 & \textbf{0.1671 }  & 0.1676 & 0.1864 & \textbf{0.1656 }  & 0.1840 & \textbf{0.1646 }  \\
\multirow{6}{*}{\begin{tabular}[c]{@{}c@{}}2\\~\\~\\~\\~\\~\\~\\ \end{tabular}} & Acc & 0.7252 & 0.7996 & 0.7718 & 0.7539 & 0.7234 & 0.7410 & 0.7234 & 0.7392 & 0.6819 & 0.7666 & \textbf{0.7963 }  & 0.7785 & 0.7523 & \textbf{0.7943 }  & 0.6860 & \textbf{0.7956 }  \\
 & AUC & 0.6862 & 0.6872 & 0.6305 & 0.6413 & \textbf{0.6824 }  & 0.6468 & \textbf{0.6824 }  & 0.6613 & 0.6481 & 0.6662 & 0.6606 & 0.6682 & 0.6672 & \textbf{0.6824 }  & 0.6495 & 0.6799 \\
 & F1 & 0.3356 & 0.2797 & 0.2865 & 0.2966 & \textbf{0.3388 }  & 0.3004 & \textbf{0.3388 }  & 0.3023 & 0.2894 & 0.2854 & 0.2960 & \textbf{0.3043 }  & 0.2965 & 0.2926 & 0.2882 & 0.2923 \\
 & G-mean & 0.6002 & 0.4839 & 0.5131 & 0.5369 & \textbf{0.6049 }  & 0.5492 & \textbf{0.6049 }  & 0.5525 & \textbf{0.5624 }  & 0.5153 & 0.5053 & 0.5291 & 0.5378 & 0.5030 & 0.5595 & 0.5017 \\
 & H\_measure & 0.1070 & 0.1097 & 0.0705 & 0.0765 & \textbf{0.1047 }  & 0.0772 & \textbf{0.1047 }  & 0.0867 & 0.0786 & 0.0915 & 0.0927 & 0.0949 & 0.0908 & \textbf{0.1056 }  & 0.0787 & 0.1038 \\
 & Brier score & 0.2489 & 0.1641 & 0.2252 & 0.2333 & 0.2547 & 0.2323 & 0.2547 & 0.2327 & 0.2820 & 0.2012 & \textbf{0.1818 }  & 0.2005 & 0.2256 & \textbf{0.1726 }  & 0.2766 & \textbf{0.1735 }  \\
\multirow{6}{*}{\begin{tabular}[c]{@{}c@{}}3\\~\\~\\~\\~\\~\\~\\ \end{tabular}} & Acc & 0.7228 & 0.7956 & 0.7746 & 0.7946 & 0.7183 & 0.7845 & 0.7183 & 0.7807 & 0.7236 & \textbf{0.8125 }  & \textbf{0.8214 }  & 0.8008 & 0.7965 & 0.8015 & 0.7268 & \textbf{0.8023 }  \\
 & AUC & 0.6862 & 0.6870 & 0.6371 & 0.6382 & \textbf{0.6793 }  & 0.6447 & \textbf{0.6793 }  & 0.6375 & 0.6397 & 0.6748 & 0.6573 & 0.6724 & 0.6673 & 0.6793 & 0.6557 & \textbf{0.6824 }  \\
 & F1 & 0.3374 & 0.2833 & \textbf{0.3022 }  & 0.2725 & \textbf{0.3383 }  & 0.2730 & \textbf{0.3383 }  & 0.2717 & 0.2881 & 0.2640 & 0.2318 & 0.2776 & 0.2670 & 0.2831 & 0.2867 & 0.2853 \\
 & G-mean & 0.6035 & 0.4914 & 0.5294 & 0.4799 & \textbf{0.6069 }  & 0.4881 & \textbf{0.6069 }  & 0.4894 & \textbf{0.5436 }  & 0.4552 & 0.4116 & 0.4807 & 0.4720 & 0.4863 & 0.5404 & 0.4881 \\
 & H\_measure & 0.1066 & 0.1091 & 0.0777 & 0.0740 & \textbf{0.1026 }  & 0.0762 & \textbf{0.1026 }  & 0.0757 & 0.0747 & 0.0996 & 0.0860 & 0.0962 & 0.0907 & \textbf{0.1027 }  & 0.0816 & \textbf{0.1059 }  \\
 & Brier score & 0.2504 & 0.1720 & 0.2220 & 0.1965 & 0.2617 & 0.1942 & 0.2617 & 0.1950 & 0.2439 & \textbf{0.1665 }  & \textbf{0.1672 }  & 0.1815 & 0.1884 & 0.1727 & 0.2404 & \textbf{0.1725 }  \\
\multirow{6}{*}{\begin{tabular}[c]{@{}c@{}}4\\~\\~\\~\\~\\~\\~\\ \end{tabular}} & Acc & 0.7369 & 0.7944 & 0.7594 & 0.8062 & 0.7188 & 0.7984 & 0.7188 & 0.7967 & 0.7432 & \textbf{0.8167 }  & \textbf{0.8103 }  & 0.8014 & \textbf{0.8080 }  & 0.8022 & 0.7448 & 0.8014 \\
 & AUC & 0.6861 & 0.6874 & 0.6387 & 0.6387 & 0.6695 & 0.6439 & 0.6695 & 0.6361 & 0.6398 & 0.6661 & 0.6537 & \textbf{0.6715 }  & 0.6655 & \textbf{0.6778 }  & 0.6594 & \textbf{0.6793 }  \\
 & F1 & 0.3296 & 0.2847 & \textbf{0.3010 }  & 0.2596 & \textbf{0.3394 }  & 0.2664 & \textbf{0.3394 }  & 0.2681 & 0.2858 & 0.2456 & 0.2707 & 0.2865 & 0.2563 & 0.2804 & 0.2829 & 0.2834 \\
 & G-mean & 0.5866 & 0.4939 & \textbf{0.5386 }  & 0.4556 & \textbf{0.6080 }  & 0.4698 & \textbf{0.6080 }  & 0.4731 & 0.5304 & 0.4311 & 0.4647 & 0.4903 & 0.4504 & 0.4825 & 0.5260 & 0.4867 \\
 & H\_measure & 0.1065 & 0.1097 & 0.0753 & 0.0756 & \textbf{0.0991 }  & 0.0776 & \textbf{0.0991 }  & 0.0776 & 0.0768 & 0.0934 & 0.0851 & 0.0973 & 0.0913 & \textbf{0.1022 }  & 0.0850 & \textbf{0.1033 }  \\
 & Brier score & 0.2399 & 0.1745 & 0.2368 & 0.1865 & 0.2608 & 0.1853 & 0.2608 & 0.1852 & 0.2271 & \textbf{0.1665 }  & 0.1733 & 0.1793 & 0.1791 & \textbf{0.1728 }  & 0.2272 & \textbf{0.1730 }  \\
\multirow{6}{*}{\begin{tabular}[c]{@{}c@{}}5\\~\\~\\~\\~\\~\\~\\ \end{tabular}} & Acc & 0.7290 & 0.7121 & 0.7041 & \textbf{0.8074 }  & 0.8034 & 0.8005 & 0.8034 & 0.8036 & 0.7444 & \textbf{0.8144 }  & 0.8007 & 0.7940 & \textbf{0.8095 }  & 0.7728 & 0.7446 & 0.7681 \\
 & AUC & 0.6859 & 0.6876 & 0.6513 & 0.6393 & 0.6674 & 0.6442 & 0.6674 & 0.6642 & 0.6550 & 0.6449 & \textbf{0.6843 }  & 0.6790 & 0.6461 & \textbf{0.6830 }  & 0.6439 & \textbf{0.6850 }  \\
 & F1 & 0.3336 & 0.3370 & \textbf{0.3192 }  & 0.2635 & 0.2808 & 0.2654 & 0.2808 & 0.2626 & 0.2810 & 0.2504 & 0.2792 & 0.2916 & 0.2600 & \textbf{0.3173 }  & 0.2810 & \textbf{0.3187 }  \\
 & G-mean & 0.5957 & 0.6082 & \textbf{0.5903 }  & 0.4590 & 0.4821 & 0.4669 & 0.4821 & 0.4612 & 0.5239 & 0.4383 & 0.4825 & 0.5022 & 0.4533 & \textbf{0.5485 }  & 0.5238 & \textbf{0.5535 }  \\
 & H\_measure & 0.1065 & 0.1102 & 0.0805 & 0.0757 & 0.0930 & 0.0768 & 0.0930 & 0.0894 & 0.0821 & 0.0841 & \textbf{0.1074 }  & 0.1034 & 0.0815 & \textbf{0.1063 }  & 0.0778 & \textbf{0.1081 }  \\
 & Brier score & 0.2461 & 0.2218 & 0.2907 & 0.1863 & 0.1922 & 0.1856 & 0.1922 & 0.1853 & 0.2308 & \textbf{0.1697 }  & \textbf{0.1792 }  & 0.1860 & \textbf{0.1774 }  & 0.1876 & 0.2273 & 0.1883 \\
\multirow{6}{*}{\begin{tabular}[c]{@{}c@{}}5.8\\~\\~\\~\\~\\~\\~\\ \end{tabular}} & Acc & 0.7422 & 0.7728 & 0.7779 & \textbf{0.8184 }  & 0.8094 & 0.8136 & 0.8094 & 0.8138 & 0.7658 & \textbf{0.8254 }  & 0.8096 & 0.8070 & \textbf{0.8199 }  & 0.8051 & 0.7663 & 0.8030 \\
 & AUC & 0.6869 & 0.6879 & 0.6438 & 0.6378 & 0.6516 & 0.6470 & 0.6516 & 0.6521 & 0.6486 & 0.6685 & 0.6515 & \textbf{0.6723 }  & 0.6670 & \textbf{0.6777 }  & 0.6619 & \textbf{0.6787 }  \\
 & F1 & 0.3299 & 0.3171 & \textbf{0.2989 }  & 0.2379 & 0.2727 & 0.2496 & 0.2727 & 0.2467 & 0.2738 & 0.2264 & 0.2701 & 0.2765 & 0.2326 & \textbf{0.2809 }  & 0.2720 & \textbf{0.2829 }  \\
 & G-mean & 0.5838 & 0.5483 & \textbf{0.5232 }  & 0.4210 & 0.4677 & 0.4382 & 0.4677 & 0.4348 & \textbf{0.5021 }  & 0.4021 & 0.4647 & 0.4741 & 0.4138 & 0.4807 & \textbf{0.4997 }  & 0.4848 \\
 & H\_measure & 0.1074 & 0.1105 & 0.0810 & 0.0751 & 0.0851 & 0.0798 & 0.0851 & 0.0830 & 0.0789 & 0.0952 & 0.0855 & \textbf{0.0981 }  & 0.0929 & \textbf{0.1035 }  & 0.0880 & \textbf{0.1040 }  \\
 & Brier score & 0.2345 & 0.1816 & 0.2186 & 0.1746 & 0.1875 & 0.1727 & 0.1875 & 0.1741 & 0.2117 & \textbf{0.1589 }  & 0.1766 & 0.1760 & \textbf{0.1667 }  & \textbf{0.1713 }  & 0.2083 & 0.1717 \\
\bottomrule
\end{tabular}
}
\end{sidewaystable}

\begin{sidewaystable}
\centering
\caption{RF results (Number of base estimators=150)}
\label{rf_results}
\resizebox{\linewidth}{!}{%
\begin{tabular}{clccccccccccccccc} 
\toprule
\multirow{2}{*}{Imbalance Ratio} & \multicolumn{1}{c}{\multirow{2}{*}{Evaluation measure}} & \multicolumn{15}{c}{Classification technique} \\ 
\cline{3-17}
 & \multicolumn{1}{c}{} & RF\_indv & RF\_posteriori & RF\_priori & RF\_lca & RF\_mcb & RF\_mla & RF\_ola & RF\_rank & RF\_metades & RF\_descluster & RF\_desknn & RF\_desp & RF\_knop & RF\_kne & RF\_knu \\ 
\hline
\multirow{6}{*}{\begin{tabular}[c]{@{}c@{}}1\\~\\~\\~\\~\\~\\~\\ \end{tabular}} & Acc & 0.6233 & 0.5767 & 0.5617 & 0.5676 & 0.5593 & 0.5676 & 0.5618 & 0.5629 & 0.6172 & \textbf{0.6295 }  & 0.6166 & \textbf{0.6238 }  & 0.6175 & \textbf{0.6263 }  & 0.6196 \\
 & AUC & 0.6934 & 0.5827 & 0.5565 & 0.5807 & 0.5558 & 0.5807 & 0.5592 & 0.5601 & 0.6832 & 0.6842 & 0.6824 & \textbf{0.6891 }  & \textbf{0.6898 }  & 0.5792 & \textbf{0.6941 }  \\
 & F1 & 0.3435 & 0.2931 & 0.2710 & 0.2914 & 0.2706 & 0.2914 & 0.2734 & 0.2741 & 0.3362 & 0.3390 & 0.3354 & \textbf{0.3421 }  & \textbf{0.3419 }  & 0.2867 & \textbf{0.3434 }  \\
 & G-mean & 0.6397 & 0.5827 & 0.5564 & 0.5804 & 0.5558 & 0.5804 & 0.5592 & 0.5601 & 0.6317 & 0.6339 & 0.6309 & \textbf{0.6381 }  & \textbf{0.6383 }  & 0.5724 & \textbf{0.6399 }  \\
 & H\_measure & 0.1151 & 0.0281 & 0.0132 & 0.0267 & 0.0128 & 0.0267 & 0.0144 & 0.0149 & 0.1049 & 0.1038 & 0.1022 & \textbf{0.1108 }  & \textbf{0.1109 }  & 0.0305 & \textbf{0.1177 }  \\
 & Brier score & 0.2227 & 0.4233 & 0.4383 & 0.4324 & 0.4407 & 0.4324 & 0.4382 & 0.4371 & 0.2317 & \textbf{0.2258 }  & \textbf{0.2256 }  & \textbf{0.2237 }  & 0.2581 & 0.3476 & 0.2371 \\
\multirow{6}{*}{\begin{tabular}[c]{@{}c@{}}2\\~\\~\\~\\~\\~\\~\\~\\ \end{tabular}} & Acc & 0.8017 & 0.7138 & 0.6539 & 0.7149 & 0.6538 & 0.7149 & 0.6523 & 0.6520 & \textbf{0.7989 }  & 0.7966 & 0.7876 & 0.7975 & \textbf{0.8038 }  & 0.7339 & \textbf{0.7992 }  \\
 & AUC & 0.6908 & 0.5711 & 0.5547 & 0.5731 & 0.5515 & 0.5731 & 0.5516 & 0.5530 & 0.6857 & 0.6808 & 0.6801 & \textbf{0.6881 }  & \textbf{0.6879 }  & 0.5691 & \textbf{0.6919 }  \\
 & F1 & 0.3028 & 0.2763 & 0.2618 & 0.2788 & 0.2582 & 0.2788 & 0.2585 & 0.2601 & 0.2814 & 0.2950 & \textbf{0.2992 }  & \textbf{0.3000 }  & 0.2907 & 0.2592 & \textbf{0.3056 }  \\
 & G-mean & 0.5085 & 0.5338 & \textbf{0.5364 }  & \textbf{0.5364 }  & 0.5320 & \textbf{0.5364 }  & 0.5327 & 0.5348 & 0.4866 & 0.5040 & 0.5161 & 0.5089 & 0.4929 & 0.5032 & 0.5139 \\
 & H\_measure & 0.1138 & 0.0274 & 0.0140 & 0.0289 & 0.0125 & 0.0289 & 0.0125 & 0.0131 & 0.1072 & 0.1026 & 0.1012 & \textbf{0.1111 }  & \textbf{0.1114 }  & 0.0293 & \textbf{0.1159 }  \\
 & Brier score & 0.1535 & 0.2862 & 0.3461 & 0.2851 & 0.3462 & 0.2851 & 0.3477 & 0.3480 & 0.1493 & 0.1563 & 0.1561 & 0.1538 & \textbf{0.1423 }  & 0.2527 & \textbf{0.1422 }  \\
\multirow{6}{*}{\begin{tabular}[c]{@{}c@{}}3\\~\\~\\~\\~\\~\\~\\~\\ \end{tabular}} & Acc & 0.8370 & 0.7677 & 0.7002 & 0.7691 & 0.6991 & 0.7691 & 0.7008 & 0.6986 & \textbf{0.8394 }  & 0.8347 & 0.8301 & 0.8360 & \textbf{0.8408 }  & 0.7813 & \textbf{0.8366 }  \\
 & AUC & 0.6892 & 0.5498 & 0.5504 & 0.5542 & 0.5497 & 0.5542 & 0.5504 & 0.5471 & \textbf{0.6885 }  & 0.6785 & 0.6759 & \textbf{0.6860 }  & 0.6848 & 0.5568 & \textbf{0.6917 }  \\
 & F1 & 0.1785 & 0.2346 & \textbf{0.2503 }  & 0.2421 & \textbf{0.2496 }  & 0.2421 & \textbf{0.2502 }  & 0.2461 & 0.1606 & 0.1899 & 0.2044 & 0.1909 & 0.1617 & 0.2300 & 0.1885 \\
 & G-mean & 0.3389 & 0.4542 & \textbf{0.5075 }  & 0.4623 & \textbf{0.5070 }  & 0.4623 & \textbf{0.5071 }  & 0.5028 & 0.3166 & 0.3535 & 0.3736 & 0.3536 & 0.3166 & 0.4399 & 0.3504 \\
 & H\_measure & 0.1119 & 0.0190 & 0.0137 & 0.0223 & 0.0133 & 0.0223 & 0.0137 & 0.0120 & \textbf{0.1110 }  & 0.0996 & 0.0974 & \textbf{0.1088 }  & 0.1083 & 0.0307 & \textbf{0.1152 }  \\
 & Brier score & 0.1325 & 0.2323 & 0.2998 & 0.2309 & 0.3009 & 0.2309 & 0.2992 & 0.3014 & \textbf{0.1306 }  & 0.1347 & 0.1349 & 0.1326 & \textbf{0.1288 }  & 0.2062 & \textbf{0.1255 }  \\
\multirow{6}{*}{\begin{tabular}[c]{@{}c@{}}4\\~\\~\\~\\~\\~\\~\\~\\ \end{tabular}} & Acc & 0.8485 & 0.8002 & 0.7262 & 0.8005 & 0.7258 & 0.8005 & 0.7268 & 0.7273 & \textbf{0.8492 }  & 0.8453 & 0.8444 & 0.8472 & \textbf{0.8496 }  & 0.8050 & \textbf{0.8481 }  \\
 & AUC & 0.6867 & 0.5434 & 0.5422 & 0.5425 & 0.5407 & 0.5425 & 0.5413 & 0.5434 & \textbf{0.6867 }  & 0.6761 & 0.6732 & \textbf{0.6849 }  & 0.6815 & 0.5580 & \textbf{0.6898 }  \\
 & F1 & 0.0992 & 0.2094 & \textbf{0.2332 }  & 0.2071 & 0.2309 & 0.2071 & \textbf{0.2316 }  & \textbf{0.2348 }  & 0.0917 & 0.1089 & 0.1239 & 0.1027 & 0.0750 & 0.1885 & 0.1028 \\
 & G-mean & 0.2355 & 0.4025 & \textbf{0.4750 }  & 0.3996 & 0.4723 & 0.3996 & \textbf{0.4726 }  & \textbf{0.4764 }  & 0.2252 & 0.2500 & 0.2694 & 0.2408 & 0.2017 & 0.3744 & 0.2404 \\
 & H\_measure & 0.1101 & 0.0200 & 0.0110 & 0.0193 & 0.0102 & 0.0193 & 0.0105 & 0.0116 & \textbf{0.1104 }  & 0.0977 & 0.0951 & \textbf{0.1082 }  & 0.1060 & 0.0302 & \textbf{0.1141 }  \\
 & Brier score & 0.1242 & 0.1998 & 0.2738 & 0.1995 & 0.2742 & 0.1995 & 0.2732 & 0.2727 & \textbf{0.1238 }  & 0.1262 & 0.1263 & \textbf{0.1243 }  & 0.1302 & 0.1805 & \textbf{0.1239 }  \\
\multirow{6}{*}{\begin{tabular}[c]{@{}c@{}}5\\~\\~\\~\\~\\~\\~\\~\\ \end{tabular}} & Acc & 0.8508 & 0.8148 & 0.7467 & 0.8166 & 0.7451 & 0.8166 & 0.7471 & 0.7460 & \textbf{0.8508 }  & 0.8493 & 0.8486 & 0.8506 & \textbf{0.8511 }  & 0.8193 & \textbf{0.8508 }  \\
 & AUC & 0.6851 & 0.5333 & 0.5410 & 0.5378 & 0.5394 & 0.5378 & 0.5376 & 0.5371 & \textbf{0.6851 }  & 0.6717 & 0.6710 & \textbf{0.6827 }  & 0.6791 & 0.5643 & \textbf{0.6888 }  \\
 & F1 & 0.0462 & 0.1758 & \textbf{0.2254 }  & 0.1862 & \textbf{0.2233 }  & 0.1862 & \textbf{0.2196 }  & 0.2191 & 0.0452 & 0.0582 & 0.0678 & 0.0549 & 0.0308 & 0.1653 & 0.0500 \\
 & G-mean & 0.1556 & 0.3525 & \textbf{0.4550 }  & 0.3635 & \textbf{0.4533 }  & 0.3635 & \textbf{0.4476 }  & 0.4475 & 0.1539 & 0.1764 & 0.1916 & 0.1705 & 0.1261 & 0.3368 & 0.1622 \\
 & H\_measure & 0.1085 & 0.0156 & 0.0116 & 0.0196 & 0.0107 & 0.0196 & 0.0099 & 0.0096 & \textbf{0.1086 }  & 0.0936 & 0.0925 & \textbf{0.1066 }  & 0.1027 & 0.0342 & \textbf{0.1129 }  \\
 & Brier score & 0.1209 & 0.1852 & 0.2533 & 0.1834 & 0.2549 & 0.1834 & 0.2529 & 0.2540 & 0.1209 & 0.1229 & 0.1230 & 0.1212 & 0.1339 & 0.1648 & 0.1261 \\
\multirow{6}{*}{\begin{tabular}[c]{@{}c@{}}5.8\\~\\~\\~\\~\\~\\~\\~\\ \end{tabular}} & Acc & 0.8519 & 0.8262 & 0.7572 & 0.8237 & 0.7567 & 0.8237 & 0.7566 & 0.7563 & \textbf{0.8520 }  & 0.8505 & 0.8509 & 0.8518 & \textbf{0.8519 }  & 0.8280 & \textbf{0.8519 }  \\
 & AUC & 0.6829 & 0.5267 & 0.5386 & 0.5267 & 0.5358 & 0.5267 & 0.5351 & 0.5356 & \textbf{0.6829 }  & 0.6701 & 0.6698 & \textbf{0.6813 }  & 0.6759 & 0.5660 & \textbf{0.6871 }  \\
 & F1 & 0.0369 & 0.1469 & \textbf{0.2177 }  & 0.1494 & \textbf{0.2129 }  & 0.1494 & 0.2116 & \textbf{0.2127 }  & 0.0370 & 0.0454 & 0.0540 & 0.0411 & 0.0216 & 0.1496 & 0.0384 \\
 & G-mean & 0.1382 & 0.3100 & \textbf{0.4398 }  & 0.3147 & \textbf{0.4341 }  & 0.3147 & 0.4326 & \textbf{0.4341 }  & 0.1382 & 0.1543 & 0.1690 & 0.1460 & 0.1050 & 0.3120 & 0.1410 \\
 & H\_measure & 0.1071 & 0.0133 & 0.0111 & 0.0126 & 0.0096 & 0.0126 & 0.0093 & 0.0095 & \textbf{0.1072 }  & 0.0943 & 0.0940 & \textbf{0.1059 }  & 0.1015 & 0.0360 & \textbf{0.1120 }  \\
 & Brier score & 0.1198 & 0.1738 & 0.2428 & 0.1763 & 0.2433 & 0.1763 & 0.2434 & 0.2437 & \textbf{0.1198 }  & \textbf{0.1214 }  & 0.1215 & \textbf{0.1200 }  & 0.1360 & 0.1562 & 0.1284 \\
\bottomrule
\end{tabular}
}
\end{sidewaystable}

\begin{sidewaystable}
\centering
\caption{MLP results (Number of base estimators=80)}
\label{MLP_results}
\resizebox{\linewidth}{!}{%
\begin{tabular}{clcccccccccccccccc} 
\cmidrule[\heavyrulewidth]{1-18}
\multirow{2}{*}{Imbalance Ratio} & \multirow{2}{*}{Evaluation measure} & \multicolumn{16}{c}{Classification  technique} \\ 
\cline{3-18}
 &  & \multicolumn{1}{l}{MLP\_indv} & \multicolumn{1}{l}{MLP\_bag} & \multicolumn{1}{l}{MLP\_posteriori} & \multicolumn{1}{l}{MLP\_priori} & \multicolumn{1}{l}{MLP\_lca} & \multicolumn{1}{l}{MLP\_mcb} & \multicolumn{1}{l}{MLP\_mla} & \multicolumn{1}{l}{MLP\_ola} & \multicolumn{1}{l}{MLP\_rank} & \multicolumn{1}{l}{MLP\_metades} & \multicolumn{1}{l}{MLP\_descluster} & \multicolumn{1}{l}{MLP\_desknn} & \multicolumn{1}{l}{MLP\_desp} & \multicolumn{1}{l}{MLP\_knop} & \multicolumn{1}{l}{MLP\_kne} & \multicolumn{1}{l}{MLP\_knu} \\ 
\hline
\multirow{6}{*}{\begin{tabular}[c]{@{}c@{}}1\\~\\~\\~\\~\\~\\ \end{tabular}} & Acc & 0.6108 & 0.6225 & 0.6065 & 0.6022 & 0.6000 & 0.5846 & 0.6000 & 0.5870 & 0.5831 & 0.6229 & 0.6220 & \textbf{0.6271}  & \textbf{0.6327}  & \textbf{0.6251}  & 0.6195 & 0.6235 \\
 & AUC & 0.6367 & 0.6758 & 0.6465 & 0.6378 & 0.6321 & 0.6153 & 0.6321 & 0.6248 & 0.6184 & 0.6714 & \textbf{0.6760}  & \textbf{0.6778}  & 0.6759 & 0.6744 & 0.6310 & \textbf{0.6791}  \\
 & F1 & 0.3091 & 0.3346 & 0.3204 & 0.3082 & 0.3099 & 0.2902 & 0.3099 & 0.2969 & 0.2917 & 0.3291 & 0.3323 & \textbf{0.3330}  & 0.3295 & \textbf{0.3335}  & 0.2978 & \textbf{0.3350}  \\
 & G-mean & 0.6008 & 0.6296 & 0.6140 & 0.6001 & 0.6022 & 0.5795 & 0.6022 & 0.5873 & 0.5813 & 0.6231 & 0.6268 & \textbf{0.6272}  & 0.6224 & \textbf{0.6280}  & 0.5866 & \textbf{0.6298}  \\
 & H\_measure & 0.0617 & 0.0965 & 0.0693 & 0.0621 & 0.0576 & 0.0459 & 0.0576 & 0.0528 & 0.0491 & 0.0922 & 0.0969 & \textbf{0.0974}  & \textbf{0.0989}  & 0.0947 & 0.0562 & \textbf{0.1002}  \\
 & Brier score & 0.2493 & 0.2303 & 0.3049 & 0.2849 & 0.2963 & 0.2841 & 0.2963 & 0.2838 & 0.2837 & \textbf{0.2310}  & \textbf{0.2315}  & \textbf{0.2312}  & 0.2326 & 0.2566 & 0.2650 & 0.2531 \\
\multirow{6}{*}{\begin{tabular}[c]{@{}c@{}}2\\~\\~\\~\\~\\~\\ \end{tabular}} & Acc & 0.7569 & 0.7845 & 0.7820 & 0.7435 & 0.7674 & 0.7024 & 0.7674 & 0.7078 & 0.6996 & 0.7848 & \textbf{0.7849}  & 0.7846 & 0.7825 & \textbf{0.7873}  & 0.7350 & \textbf{0.7856}  \\
 & AUC & 0.6626 & 0.6812 & 0.6328 & 0.6379 & 0.6290 & 0.6219 & 0.6290 & 0.6216 & 0.6198 & \textbf{0.6806}  & 0.6787 & \textbf{0.6804}  & \textbf{0.6789}  & 0.6692 & 0.6361 & 0.6778 \\
 & F1 & 0.2962 & 0.2974 & 0.2881 & 0.2904 & 0.2915 & 0.2788 & 0.2915 & 0.2764 & 0.2758 & \textbf{0.2937}  & \textbf{0.2948}  & 0.2909 & 0.2928 & 0.2934 & 0.2675 & \textbf{0.2964}  \\
 & G-mean & 0.5345 & 0.5164 & 0.5075 & 0.5358 & 0.5220 & \textbf{0.5418}  & 0.5220 & \textbf{0.5366}  & \textbf{0.5393}  & 0.5120 & 0.5132 & 0.5088 & 0.5126 & 0.5096 & 0.5127 & 0.5144 \\
 & H\_measure & 0.0840 & 0.1032 & 0.0685 & 0.0654 & 0.0624 & 0.0501 & 0.0624 & 0.0500 & 0.0477 & \textbf{0.1026}  & 0.1006 & 0.1008 & \textbf{0.1028}  & 0.0953 & 0.0590 & \textbf{0.1022}  \\
 & Brier score & 0.1684 & 0.1561 & 0.1761 & 0.1847 & 0.1820 & 0.2022 & 0.1820 & 0.2005 & 0.2040 & 0.1561 & \textbf{0.1544}  & 0.1558 & 0.1573 & \textbf{0.1525}  & 0.1892 & \textbf{0.1521}  \\
\multirow{6}{*}{\begin{tabular}[c]{@{}c@{}}3\\~\\~\\~\\~\\~\\ \end{tabular}} & Acc & 0.8062 & 0.8293 & 0.8259 & 0.7962 & 0.8197 & 0.7604 & 0.8197 & 0.7661 & 0.7568 & 0.8296 & \textbf{0.8299}  & 0.8295 & 0.8274 & \textbf{0.8313}  & 0.7849 & \textbf{0.8310}  \\
 & AUC & 0.6544 & 0.6796 & 0.6013 & 0.6237 & 0.6118 & 0.6261 & 0.6117 & 0.6266 & 0.6266 & \textbf{0.6796}  & \textbf{0.6777}  & \textbf{0.6784}  & 0.6772 & 0.6410 & 0.6426 & 0.6576 \\
 & F1 & 0.2353 & 0.2087 & 0.2023 & 0.2304 & 0.1915 & \textbf{0.2566}  & 0.1915 & \textbf{0.2507}  & \textbf{0.2522}  & 0.2061 & 0.2018 & 0.2048 & 0.2024 & 0.1990 & 0.2371 & 0.2088 \\
 & G-mean & 0.4279 & 0.3791 & 0.3747 & 0.4299 & 0.3672 & \textbf{0.4851}  & 0.3672 & \textbf{0.4745}  & \textbf{0.4820}  & 0.3759 & 0.0994 & 0.3746 & 0.3736 & 0.3665 & 0.4459 & 0.3779 \\
 & H\_measure & 0.0773 & 0.1013 & 0.0475 & 0.0548 & 0.0481 & 0.0526 & 0.0481 & 0.0538 & 0.0536 & \textbf{0.1013}  & \textbf{0.0994}  & 0.0988 & \textbf{0.1003}  & 0.0773 & 0.0648 & 0.0889 \\
 & Brier score & 0.1437 & 0.1335 & 0.1470 & 0.1535 & 0.1480 & 0.1693 & 0.1480 & 0.1659 & 0.1698 & 0.1335 & \textbf{0.1321}  & 0.1331 & 0.1336 & \textbf{0.1327}  & 0.1585 & \textbf{0.1314}  \\
\multirow{6}{*}{\begin{tabular}[c]{@{}c@{}}4\\~\\~\\~\\~\\~\\ \end{tabular}} & Acc & 0.8338 & 0.8443 & 0.8426 & 0.8230 & 0.8344 & 0.7954 & 0.8344 & 0.7979 & 0.7931 & 0.8445 & \textbf{0.8446}  & 0.8434 & 0.8432 & \textbf{0.8460}  & 0.8143 & \textbf{0.8452}  \\
 & AUC & 0.6588 & 0.6835 & 0.5990 & 0.6256 & 0.6150 & 0.6340 & 0.6150 & 0.6346 & 0.6375 & \textbf{0.6835}  & \textbf{0.6821}  & \textbf{0.6819}  & 0.6798 & 0.6227 & 0.6576 & 0.6369 \\
 & F1 & 0.1765 & 0.1233 & 0.1244 & 0.1812 & 0.1419 & \textbf{0.2218}  & 0.1419 & \textbf{0.2200}  & \textbf{0.2260}  & 0.1242 & 0.1186 & 0.1192 & 0.1248 & 0.1135 & 0.1938 & 0.1229 \\
 & G-mean & 0.3392 & 0.2687 & 0.2713 & 0.3529 & 0.2982 & \textbf{0.4205}  & 0.2982 & \textbf{0.4166}  & \textbf{0.4270}  & 0.2697 & 0.2627 & 0.2642 & 0.2714 & 0.2555 & 0.3739 & 0.2677 \\
 & H\_measure & 0.0822 & 0.1058 & 0.0438 & 0.0575 & 0.0491 & 0.0594 & 0.0491 & 0.0603 & 0.0611 & \textbf{0.1058}  & \textbf{0.1039}  & 0.1016 & \textbf{0.1036}  & 0.0690 & 0.0767 & 0.0784 \\
 & Brier score & 0.1305 & 0.1238 & 0.1349 & 0.1387 & 0.1380 & 0.1496 & 0.1380 & 0.1484 & 0.1505 & \textbf{0.1238}  & \textbf{0.1231}  & \textbf{0.1238}  & 0.1242 & 0.1284 & 0.1407 & 0.1273 \\
\multirow{6}{*}{\begin{tabular}[c]{@{}c@{}}5\\~\\~\\~\\~\\~\\ \end{tabular}} & Acc & 0.8403 & 0.8484 & 0.8472 & 0.8352 & 0.8445 & 0.8140 & 0.8445 & 0.8149 & 0.8105 & \textbf{0.8483}  & 0.8480 & 0.8480 & 0.8482 & \textbf{0.8489}  & 0.8261 & \textbf{0.8488}  \\
 & AUC & 0.6591 & 0.6849 & 0.5939 & 0.6203 & 0.6167 & 0.6386 & 0.6168 & 0.6407 & 0.6447 & \textbf{0.6849}  & \textbf{0.6833}  & \textbf{0.6827}  & 0.6818 & 0.5968 & 0.6620 & 0.6141 \\
 & F1 & 0.1323 & 0.0891 & 0.0781 & 0.1368 & 0.0988 & \textbf{0.1920}  & 0.0988 & \textbf{0.1869}  & \textbf{0.2040}  & 0.0867 & 0.0813 & 0.0765 & 0.0836 & 0.0789 & 0.1691 & 0.0810 \\
 & G-mean & 0.2824 & 0.2222 & 0.2075 & 0.2915 & 0.2374 & \textbf{0.3721}  & 0.2374 & \textbf{0.3656}  & \textbf{0.3887}  & 0.2189 & 0.2116 & 0.2047 & 0.2147 & 0.2077 & 0.3365 & 0.2108 \\
 & H\_measure & 0.0824 & 0.1075 & 0.0380 & 0.0549 & 0.0505 & 0.0635 & 0.0505 & 0.0645 & 0.0680 & \textbf{0.1075}  & \textbf{0.1056}  & 0.1037 & \textbf{0.1048}  & 0.0569 & 0.0815 & 0.0634 \\
 & Brier score & 0.1268 & 0.1210 & 0.1323 & 0.1333 & 0.1320 & 0.1408 & 0.1320 & 0.1401 & 0.1416 & \textbf{0.1210}  & \textbf{0.1209}  & \textbf{0.1212}  & 0.1214 & 0.1293 & 0.1340 & 0.1283 \\
\multirow{6}{*}{\begin{tabular}[c]{@{}c@{}}5.8\\~\\~\\~\\~\\~\\ \end{tabular}} & Acc & 0.8398 & 0.8494 & 0.8490 & 0.8391 & 0.8462 & 0.8234 & 0.8462 & 0.8226 & 0.8181 & 0.8494 & \textbf{0.8498}  & 0.8491 & 0.8492 & \textbf{0.8499}  & 0.8305 & \textbf{0.8499}  \\
 & AUC & 0.6607 & 0.6837 & 0.5995 & 0.6230 & 0.6090 & 0.6450 & 0.6090 & 0.6391 & 0.6412 & \textbf{0.6837}  & \textbf{0.6811}  & \textbf{0.6823}  & 0.6805 & 0.5894 & 0.6653 & 0.6044 \\
 & F1 & 0.1347 & 0.0656 & 0.0478 & 0.1135 & 0.0657 & \textbf{0.1850}  & 0.0657 & \textbf{0.1756}  & \textbf{0.1808}  & 0.0642 & 0.0630 & 0.0620 & 0.0694 & 0.0596 & 0.1438 & 0.0631 \\
 & G-mean & 0.2857 & 0.1879 & 0.1592 & 0.2599 & 0.1896 & \textbf{0.3569}  & 0.1896 & \textbf{0.3467}  & \textbf{0.3560}  & 0.1858 & 0.1837 & 0.1826 & 0.1937 & 0.1783 & 0.3033 & 0.1838 \\
 & H\_measure & 0.0826 & 0.1062 & 0.0399 & 0.0542 & 0.0436 & 0.0699 & 0.0436 & 0.0648 & 0.0655 & \textbf{0.1062}  & \textbf{0.1034}  & 0.1030 & \textbf{0.1046}  & 0.0490 & 0.0850 & 0.0557 \\
 & Brier score & 0.1277 & 0.1204 & 0.1306 & 0.1317 & 0.1308 & 0.1360 & 0.1308 & 0.1363 & 0.1378 & \textbf{0.1204}  & \textbf{0.1206}  & \textbf{0.1206}  & 0.1208 & 0.1300 & 0.1310 & 0.1290 \\
\bottomrule
\end{tabular}
}
\end{sidewaystable}

\begin{sidewaystable}
\centering
\caption{Heterogeneous pool of classifiers results }
\label{het_results}
\resizebox{\linewidth}{!}{%
\begin{tabular}{clccccccccccccccc} 
\toprule
\multirow{2}{*}{Imbalance Ratio} & \multicolumn{1}{c}{\multirow{2}{*}{Evaluation measure}} & \multicolumn{15}{c}{Classification technique} \\ 
\cline{3-17}
 & \multicolumn{1}{c}{} & Pool & Pool\_posteriori & Pool\_priori & Pool\_lca & Pool\_mcb & Pool\_mla & Pool\_ola & Pool\_rank & Pool\_metades & Pool\_descluster & Pool\_desknn & Pool\_desp & Pool\_knop & Pool\_kne & Pool\_knu \\ 
\hline
\multirow{6}{*}{\begin{tabular}[c]{@{}c@{}}1\\~\\~\\~\\~\\~\\~\\ \end{tabular}} & Acc & 0.7406 & \textbf{0.8137 }  & 0.6692 & 0.6330 & 0.6497 & 0.6330 & 0.6373 & 0.6478 & 0.6637 & 0.6623 & \textbf{0.7307 }  & \textbf{0.6885 }  & 0.6552 & 0.6755 & 0.6632 \\
 & AUC & 0.6856 & 0.5353 & 0.6360 & 0.6654 & 0.6344 & 0.6654 & 0.6485 & 0.6222 & 0.6623 & \textbf{0.6933 }  & 0.6713 & 0.6614 & \textbf{0.6736 }  & 0.6277 & \textbf{0.6813 }  \\
 & F1 & 0.3310 & 0.2526 & 0.3274 & 0.3389 & 0.3257 & 0.3389 & 0.3262 & 0.3133 & 0.3349 & \textbf{0.3501 }  & 0.3282 & 0.3377 & \textbf{0.3429 }  & 0.3211 & \textbf{0.3462 }  \\
 & G-mean & 0.5861 & 0.4415 & 0.6124 & 0.6333 & 0.6151 & 0.6334 & 0.6179 & 0.6007 & 0.6228 & \textbf{0.6413 }  & 0.5884 & 0.6187 & \textbf{0.6343 }  & 0.6030 & \textbf{0.6364 }  \\
 & H\_measure & 0.1094 & 0.0513 & 0.0795 & 0.0928 & 0.0748 & 0.0929 & 0.0800 & 0.0652 & 0.0880 & \textbf{0.1180 }  & 0.0972 & 0.0926 & \textbf{0.0974 }  & 0.0743 & \textbf{0.1060 }  \\
 & Brier score & 0.1813 & \textbf{0.1713 }  & 0.2152 & 0.2219 & 0.2193 & 0.2219 & 0.2217 & 0.2193 & 0.1998 & 0.2110 & \textbf{0.1941 }  & 0.1983 & 0.2011 & 0.2075 & \textbf{0.1914 }  \\
\multirow{6}{*}{\begin{tabular}[c]{@{}c@{}}2\\~\\~\\~\\~\\~\\~\\ \end{tabular}} & Acc & 0.7773 & 0.7522 & 0.7733 & 0.7975 & 0.7674 & 0.7975 & 0.7685 & 0.7555 & \textbf{0.8016 }  & \textbf{0.8248 }  & \textbf{0.8142 }  & 0.7983 & 0.7995 & 0.7720 & 0.7982 \\
 & AUC & 0.6878 & 0.6488 & 0.6598 & 0.6778 & 0.6680 & 0.6779 & 0.6720 & 0.6694 & \textbf{0.6882 }  & \textbf{0.6913 }  & 0.6812 & \textbf{0.6832 }  & 0.6733 & 0.6756 & 0.6764 \\
 & F1 & 0.3267 & \textbf{0.3288 }  & 0.3030 & 0.2951 & 0.3041 & 0.2953 & 0.2996 & 0.3005 & 0.3031 & 0.2492 & 0.2684 & 0.2968 & \textbf{0.3061 }  & 0.3016 & \textbf{0.3067 }  \\
 & G-mean & 0.5561 & \textbf{0.5763 }  & 0.5313 & 0.5033 & \textbf{0.5369 }  & 0.5035 & 0.5308 & \textbf{0.5405 }  & 0.5089 & 0.4279 & 0.4587 & 0.5046 & 0.5141 & 0.5306 & 0.5159 \\
 & H\_measure & 0.1123 & 0.0872 & 0.0900 & 0.1009 & 0.0919 & 0.1010 & 0.0937 & 0.0912 & \textbf{0.1118 }  & \textbf{0.1178 }  & \textbf{0.1056 }  & 0.1055 & 0.1032 & 0.0977 & 0.1043 \\
 & Brier score & 0.1543 & 0.2212 & 0.1759 & 0.1584 & 0.1716 & 0.1585 & 0.1655 & 0.1739 & \textbf{0.1537 }  & \textbf{0.1425 }  & 0.1596 & 0.1597 & 0.1563 & 0.1723 & \textbf{0.1555 }  \\
\multirow{6}{*}{\begin{tabular}[c]{@{}c@{}}3\\~\\~\\~\\~\\~\\~\\ \end{tabular}} & Acc & 0.8138 & 0.7610 & 0.8110 & \textbf{0.8357 }  & 0.8076 & 0.8356 & 0.8092 & 0.7928 & 0.8312 & \textbf{0.8477 }  & \textbf{0.8384 }  & 0.8278 & 0.8325 & 0.8028 & 0.8318 \\
 & AUC & 0.6837 & 0.6674 & 0.6659 & 0.6665 & 0.6684 & 0.6664 & 0.6641 & 0.6652 & \textbf{0.6833 }  & \textbf{0.6849 }  & 0.6779 & \textbf{0.6791 }  & 0.6618 & 0.6719 & 0.6698 \\
 & F1 & 0.2762 & \textbf{0.3226 }  & 0.2388 & 0.1811 & 0.2433 & 0.1810 & 0.2298 & \textbf{0.2569 }  & 0.1945 & 0.0918 & 0.1574 & 0.1995 & 0.1955 & \textbf{0.2459 }  & 0.2013 \\
 & G-mean & 0.4678 & \textbf{0.5631 }  & 0.4282 & 0.3428 & 0.4360 & 0.3428 & 0.4194 & \textbf{0.4632 }  & 0.3616 & 0.2261 & 0.3136 & 0.3701 & 0.3617 & \textbf{0.4428 }  & 0.3687 \\
 & H\_measure & 0.1085 & 0.0952 & 0.0918 & 0.0905 & 0.0918 & 0.0904 & 0.0865 & 0.0867 & \textbf{0.1078 }  & \textbf{0.1098 }  & 0.1008 & \textbf{0.1010 }  & 0.0919 & 0.0932 & 0.0989 \\
 & Brier score & 0.1364 & 0.2202 & 0.1514 & \textbf{0.1332 }  & 0.1515 & \textbf{0.1333 }  & 0.1460 & 0.1603 & 0.1355 & \textbf{0.1250 }  & 0.1395 & 0.1424 & 0.1347 & 0.1594 & 0.1337 \\
\multirow{6}{*}{\begin{tabular}[c]{@{}c@{}}4\\~\\~\\~\\~\\~\\~\\ \end{tabular}} & Acc & 0.8354 & 0.7806 & 0.8310 & \textbf{0.8474 }  & 0.8256 & \textbf{0.8474 }  & 0.8284 & 0.8125 & 0.8456 & \textbf{0.8512 }  & 0.8469 & 0.8388 & 0.8456 & 0.8172 & 0.8454 \\
 & AUC & 0.6820 & 0.6663 & 0.6594 & 0.6612 & 0.6654 & 0.6613 & 0.6643 & 0.6640 & \textbf{0.6819 }  & \textbf{0.6839 }  & 0.6758 & \textbf{0.6785 }  & 0.6506 & 0.6741 & 0.6574 \\
 & F1 & 0.1944 & \textbf{0.3079 }  & 0.1812 & 0.0836 & 0.1838 & 0.0836 & 0.1771 & \textbf{0.2136 }  & 0.1267 & 0.0273 & 0.0812 & 0.1347 & 0.1179 & \textbf{0.1957 }  & 0.1211 \\
 & G-mean & 0.3580 & \textbf{0.5316 }  & 0.3467 & 0.2152 & 0.3540 & 0.2152 & 0.3440 & \textbf{0.3983 }  & 0.2720 & 0.1185 & 0.2121 & 0.2864 & 0.2611 & \textbf{0.3741 }  & 0.2653 \\
 & H\_measure & 0.1072 & 0.0948 & 0.0863 & 0.0855 & 0.0899 & 0.0856 & 0.0881 & 0.0872 & \textbf{0.1071 }  & \textbf{0.1104 }  & 0.0992 & \textbf{0.1012 }  & 0.0832 & 0.0961 & 0.0905 \\
 & Brier score & 0.1273 & 0.2022 & 0.1386 & \textbf{0.1247 }  & 0.1391 & \textbf{0.1247 }  & 0.1362 & 0.1507 & 0.1270 & \textbf{0.1199 }  & 0.1288 & 0.1330 & 0.1262 & 0.1493 & 0.1251 \\
\multirow{6}{*}{\begin{tabular}[c]{@{}c@{}}5\\~\\~\\~\\~\\~\\~\\ \end{tabular}} & Acc & 0.8438 & 0.7872 & 0.8373 & 0.8496 & 0.8356 & \textbf{0.8496 }  & 0.8369 & 0.8193 & 0.8485 & \textbf{0.8519 }  & 0.8496 & 0.8443 & 0.8492 & 0.8225 & 0.8491 \\
 & AUC & 0.6820 & 0.6681 & 0.6559 & 0.6564 & 0.6626 & 0.6564 & 0.6571 & 0.6589 & \textbf{0.6806 }  & \textbf{0.6823 }  & 0.6732 & \textbf{0.6759 }  & 0.6362 & 0.6725 & 0.6439 \\
 & F1 & 0.1423 & \textbf{0.3060 }  & 0.1242 & 0.0477 & 0.1403 & 0.0477 & 0.1253 & \textbf{0.1929 }  & 0.0729 & 0.0152 & 0.0453 & 0.0904 & 0.0689 & \textbf{0.1791 }  & 0.0680 \\
 & G-mean & 0.2919 & \textbf{0.5242 }  & 0.2746 & 0.1588 & 0.2954 & 0.1588 & 0.2762 & \textbf{0.3693 }  & 0.1993 & 0.0878 & 0.1546 & 0.2263 & 0.1930 & \textbf{0.3509 }  & 0.1917 \\
 & H\_measure & 0.1073 & 0.0958 & 0.0816 & 0.0797 & 0.0878 & 0.0797 & 0.0804 & 0.0814 & \textbf{0.1048 }  & \textbf{0.1075 }  & 0.0947 & \textbf{0.0970 }  & 0.0718 & 0.0944 & 0.0779 \\
 & Brier score & 0.1245 & 0.1965 & 0.1331 & 0.1238 & 0.1335 & 0.1238 & 0.1320 & 0.1486 & \textbf{0.1237 }  & \textbf{0.1195 }  & 0.1254 & 0.1286 & 0.1245 & 0.1467 & \textbf{0.1238 }  \\
\multirow{6}{*}{\begin{tabular}[c]{@{}c@{}}5.8\\~\\~\\~\\~\\~\\~\\ \end{tabular}} & Acc & 0.8459 & 0.7968 & 0.8415 & 0.8492 & 0.8383 & 0.8493 & 0.8399 & 0.8232 & 0.8492 & \textbf{0.8518 }  & \textbf{0.8505 }  & 0.8462 & \textbf{0.8507 }  & 0.8267 & 0.8504 \\
 & AUC & 0.6813 & 0.6653 & 0.6505 & 0.6582 & 0.6614 & 0.6582 & 0.6619 & 0.6634 & \textbf{0.6805 }  & \textbf{0.6815 }  & 0.6740 & \textbf{0.6761 }  & 0.6363 & 0.6749 & 0.6409 \\
 & F1 & 0.1112 & \textbf{0.2932 }  & 0.1035 & 0.0493 & 0.1200 & 0.0497 & 0.1246 & \textbf{0.1825 }  & 0.0602 & 0.0105 & 0.0356 & 0.0800 & 0.0533 & \textbf{0.1674 }  & 0.0558 \\
 & G-mean & 0.2527 & \textbf{0.5016 }  & 0.2455 & 0.1616 & 0.2687 & 0.1625 & 0.2734 & \textbf{0.3543 }  & 0.1797 & 0.0729 & 0.1361 & 0.2108 & 0.1678 & \textbf{0.3340 }  & 0.1720 \\
 & H\_measure & 0.1065 & 0.0935 & 0.0778 & 0.0807 & 0.0865 & 0.0808 & 0.0853 & 0.0852 & \textbf{0.1048 }  & \textbf{0.1065 }  & 0.0963 & \textbf{0.0975 }  & 0.0704 & 0.0960 & 0.0736 \\
 & Brier score & 0.1231 & 0.1868 & 0.1306 & 0.1242 & 0.1316 & 0.1242 & 0.1297 & 0.1460 & \textbf{0.1226 }  & \textbf{0.1199 }  & 0.1240 & 0.1266 & 0.1244 & 0.1441 & \textbf{0.1239 }  \\
\bottomrule
\end{tabular}
}
\end{sidewaystable}

\newpage
\bibliographystyle{apacite}
\bibliography{references}
\end{document}